\documentclass{article}

\usepackage{microtype}
\usepackage{graphicx}
\usepackage{booktabs} %

\usepackage{hyperref}

\usepackage[accepted]{configuration/icml2025}

\usepackage{amsmath}
\usepackage{amssymb}
\usepackage{mathtools}
\usepackage{amsthm}

\usepackage[capitalize,noabbrev]{cleveref}

\usepackage{amsmath,amssymb,amsthm}

\providecommand{\norm}[1]{\left\lVert#1\right\rVert}

\providecommand{\R}{\mathbb{R}} %

\providecommand{\E}{{\mathbb E}}
\providecommand{\E}[1]{{\mathbb E}\left.#1\right. }        %

\providecommand{\pp}{\mathbf{p}}

\renewcommand{\ss}{\mathbf{s}}

\providecommand{\vv}{\mathbf{v}}

\providecommand{\xx}{\mathbf{x}}

\providecommand{\zz}{\mathbf{z}}

\providecommand{\mC}{\mathbf{C}}

\providecommand{\mI}{\mathbf{I}}

\providecommand{\mM}{\mathbf{M}}

\providecommand{\mW}{\mathbf{W}}

\providecommand{\mepsilon}{\boldsymbol{\epsilon}}

\providecommand{\mpsi}{\boldsymbol{\psi}}
\providecommand{\mtheta}{\boldsymbol{\theta}}
\providecommand{\mmu}{\boldsymbol{\mu}}
\providecommand{\mpsi}{\boldsymbol{\psi}}
\providecommand{\mf}{\boldsymbol{f}}

\providecommand{\cL}{\mathcal{L}}

\providecommand{\cN}{\mathcal{N}}

\newenvironment{talign*}
{\csname align*\endcsname}
{\endalign}

\usepackage[utf8]{inputenc}         %
\usepackage[T1]{fontenc}            %
\usepackage{url}                    %
\usepackage{booktabs}               %
\usepackage{amsfonts}               %
\usepackage{nicefrac}               %
\usepackage{microtype}              %
\usepackage{xcolor}                 %
\usepackage{algpseudocode}
\usepackage{algorithm}
\usepackage{graphicx}
\usepackage{subcaption}
\usepackage[flushleft]{threeparttable}
\usepackage{float}
\usepackage{multirow}
\usepackage{xspace}
\usepackage{natbib}
\usepackage{enumitem}
\usepackage[font=small]{caption}
\usepackage{autobreak}
\usepackage{sidecap}
\usepackage{wrapfig}
\usepackage{bbding}
\usepackage[toc, page, header]{appendix}
\usepackage{tikz}
\usepackage{xcolor}
\usepackage{pifont}
\usepackage{mdframed}
\usepackage{colortbl}
\usepackage[english]{babel}

\hypersetup{
  colorlinks=true,
  linkcolor=blue,
  citecolor=blue,
  urlcolor=blue
}

\definecolor{coral}{RGB}{255,127,80}
\definecolor{darkgreen}{RGB}{0,100,0}
\definecolor{darkyellow}{RGB}{204,153,0}
\definecolor{salmon}{RGB}{250,128,114}

\newcommand{\transparentred}[1]{%
  \tikz[baseline=(X.base)] \node[fill=red, fill opacity=0.1, text opacity=1, inner sep=2pt, outer sep=0pt] (X) {#1};%
}
\newcommand{\thmref}[1]{\hyperref[#1]{\transparentred{Theorem~\ref*{#1}}}}

\newcommand{\transparentgray}[1]{%
  \tikz[baseline=(X.base)] \node[fill=gray, fill opacity=0.1, text opacity=1, inner sep=2pt, outer sep=0pt] (X) {#1};%
}
\newcommand{\defref}[1]{\hyperref[#1]{\transparentgray{Definition~\ref*{#1}}}}

\newcommand{\transparentblue}[1]{%
  \tikz[baseline=(X.base)] \node[fill=blue, fill opacity=0.1, text opacity=1, inner sep=2pt, outer sep=0pt] (X) {#1};%
}
\newcommand{\propref}[1]{\hyperref[#1]{\transparentblue{Proposition~\ref*{#1}}}}

\newcommand{\transparentgreen}[1]{%
  \tikz[baseline=(X.base)] \node[fill=green, fill opacity=0.1, text opacity=1, inner sep=2pt, outer sep=0pt] (X) {#1};%
}
\newcommand{\assumpref}[1]{\hyperref[#1]{\transparentgreen{Assumption~\ref*{#1}}}}

\newcommand{\transparentyellow}[1]{%
  \tikz[baseline=(X.base)] \node[fill=yellow, fill opacity=0.1, text opacity=1, inner sep=2pt, outer sep=0pt] (X) {#1};%
}
\newcommand{\remarkref}[1]{\hyperref[#1]{\transparentyellow{Remark~\ref*{#1}}}}

\newcommand{\transparentpurple}[1]{%
  \tikz[baseline=(X.base)] \node[fill=purple, fill opacity=0.1, text opacity=1, inner sep=2pt, outer sep=0pt] (X) {#1};%
}
\newcommand{\hypref}[1]{\hyperref[#1]{\transparentpurple{Hypothesis~\ref*{#1}}}}

\newcommand{\transparentorange}[1]{%
  \tikz[baseline=(X.base)] \node[fill=orange, fill opacity=0.1, text opacity=1, inner sep=2pt, outer sep=0pt] (X) {#1};%
}
\newcommand{\conjref}[1]{\hyperref[#1]{\transparentorange{Conjecture~\ref*{#1}}}}

\newcommand{\transparentcyan}[1]{%
  \tikz[baseline=(X.base)] \node[fill=cyan, fill opacity=0.1, text opacity=1, inner sep=2pt, outer sep=0pt] (X) {#1};%
}
\newcommand{\lemref}[1]{\hyperref[#1]{\transparentcyan{Lemma~\ref*{#1}}}}

\newcommand{\transparentmagenta}[1]{%
  \tikz[baseline=(X.base)] \node[fill=magenta, fill opacity=0.1, text opacity=1, inner sep=2pt, outer sep=0pt] (X) {#1};%
}
\newcommand{\corref}[1]{\hyperref[#1]{\transparentmagenta{Corollary~\ref*{#1}}}}

\newcommand{\transparentlime}[1]{%
  \tikz[baseline=(X.base)] \node[fill=lime, fill opacity=0.1, text opacity=1, inner sep=2pt, outer sep=0pt] (X) {#1};%
}
\newcommand{\exampleref}[1]{\hyperref[#1]{\transparentlime{Example~\ref*{#1}}}}

\newcommand{\transparentpink}[1]{%
  \tikz[baseline=(X.base)] \node[fill=pink, fill opacity=0.1, text opacity=1, inner sep=2pt, outer sep=0pt] (X) {#1};%
}
\newcommand{\noteref}[1]{\hyperref[#1]{\transparentpink{Notation~\ref*{#1}}}}

\newcommand{\transparentviolet}[1]{%
  \tikz[baseline=(X.base)] \node[fill=violet, fill opacity=0.1, text opacity=1, inner sep=2pt, outer sep=0pt] (X) {#1};%
}
\newcommand{\claimref}[1]{\hyperref[#1]{\transparentviolet{Claim~\ref*{#1}}}}

\newcommand{\transparentsalmon}[1]{%
  \tikz[baseline=(X.base)] \node[fill=salmon, fill opacity=0.1, text opacity=1, inner sep=2pt, outer sep=0pt] (X) {#1};%
}
\newcommand{\probref}[1]{\hyperref[#1]{\transparentsalmon{Problem~\ref*{#1}}}}

\newcommand{\transparentlavender}[1]{%
  \tikz[baseline=(X.base)] \node[fill=lavender, fill opacity=0.1, text opacity=1, inner sep=2pt, outer sep=0pt] (X) {#1};%
}
\newcommand{\obsref}[1]{\hyperref[#1]{\transparentlavender{Observation~\ref*{#1}}}}

\newcommand{\transparentteal}[1]{%
  \tikz[baseline=(X.base)] \node[fill=teal, fill opacity=0.1, text opacity=1, inner sep=2pt, outer sep=0pt] (X) {#1};%
}
\newcommand{\figref}[1]{\hyperref[#1]{\transparentteal{Figure~\ref*{#1}}}}

\newcommand{\transparentdarkgreen}[1]{%
  \tikz[baseline=(X.base)] \node[fill=darkgreen, fill opacity=0.1, text opacity=1, inner sep=2pt, outer sep=0pt] (X) {#1};%
}
\newcommand{\tabref}[1]{\hyperref[#1]{\transparentdarkgreen{Table~\ref*{#1}}}}

\newcommand{\transparentdarkyellow}[1]{%
  \tikz[baseline=(X.base)] \node[fill=darkyellow, fill opacity=0.1, text opacity=1, inner sep=2pt, outer sep=0pt] (X) {#1};%
}
\newcommand{\secref}[1]{\hyperref[#1]{\transparentdarkyellow{Section~\ref*{#1}}}}

\newcommand{\transparentcoral}[1]{%
  \tikz[baseline=(X.base)] \node[fill=coral, fill opacity=0.1, text opacity=1, inner sep=2pt, outer sep=0pt] (X) {#1};%
}
\newcommand{\appref}[1]{\hyperref[#1]{\transparentcoral{Appendix~\ref*{#1}}}}
\newcommand{\algoref}[1]{\hyperref[#1]{\transparentteal{Algorithm~\ref*{#1}}}}
\def\pz{{\phantom{0}}}

\newtheoremstyle{custom}
{1pt} %
{1pt} %
{\itshape} %
{} %
{\bfseries} %
{} %
{ } %
{\thmname{#1} \thmnumber{#2} \thmnote{(#3)} . } %

\theoremstyle{custom}

\newtheorem{innerdefinition}{Definition}
\newtheorem{innerproposition}{Proposition}
\newtheorem{innerassumption}{Assumption}
\newtheorem{innerremark}{Remark}
\newtheorem{innertheorem}{Theorem}
\newtheorem{innerhypothesis}{Hypothesis}
\newtheorem{innerconjecture}{Conjecture}
\newtheorem{innerlemma}{Lemma}
\newtheorem{innercorollary}{Corollary}
\newtheorem{innerexample}{Example}
\newtheorem{innernotation}{Notation}
\newtheorem{innerclaim}{Claim}
\newtheorem{innerproblem}{Problem}

\newtheorem{innerobservation}{Observation}

\newmdenv[
  backgroundcolor=gray!10,
  linecolor=gray!100,
  linewidth=0.8pt,
  skipabove=2pt,
  skipbelow=2pt,
  innertopmargin=10pt,
  innerbottommargin=5pt,
  innerleftmargin=5pt,
  innerrightmargin=5pt,
]{definitionframe}

\newmdenv[
  backgroundcolor=blue!10,
  linecolor=blue!100,
  linewidth=0.8pt,
  skipabove=2pt,
  skipbelow=2pt,
  innertopmargin=10pt,
  innerbottommargin=5pt,
  innerleftmargin=5pt,
  innerrightmargin=5pt,
]{propositionframe}

\newmdenv[
  backgroundcolor=green!10,
  linecolor=green!100,
  linewidth=0.8pt,
  skipabove=2pt,
  skipbelow=2pt,
  innertopmargin=10pt,
  innerbottommargin=5pt,
  innerleftmargin=5pt,
  innerrightmargin=5pt,
]{assumptionframe}

\newmdenv[
  backgroundcolor=yellow!10,
  linecolor=yellow!100,
  linewidth=0.8pt,
  skipabove=2pt,
  skipbelow=2pt,
  innertopmargin=10pt,
  innerbottommargin=5pt,
  innerleftmargin=5pt,
  innerrightmargin=5pt,
]{remarkframe}

\newmdenv[
  backgroundcolor=red!10,
  linecolor=red!100,
  linewidth=0.8pt,
  skipabove=2pt,
  skipbelow=2pt,
  innertopmargin=10pt,
  innerbottommargin=5pt,
  innerleftmargin=5pt,
  innerrightmargin=5pt,
]{theoremframe}

\newmdenv[
  backgroundcolor=purple!10,
  linecolor=purple!100,
  linewidth=0.8pt,
  skipabove=2pt,
  skipbelow=2pt,
  innertopmargin=10pt,
  innerbottommargin=5pt,
  innerleftmargin=5pt,
  innerrightmargin=5pt,
]{hypothesisframe}

\newmdenv[
  backgroundcolor=orange!10,
  linecolor=orange!100,
  linewidth=0.8pt,
  skipabove=2pt,
  skipbelow=2pt,
  innertopmargin=10pt,
  innerbottommargin=5pt,
  innerleftmargin=5pt,
  innerrightmargin=5pt,
]{conjectureframe}

\newmdenv[
  backgroundcolor=cyan!10,
  linecolor=cyan!100,
  linewidth=0.8pt,
  skipabove=2pt,
  skipbelow=2pt,
  innertopmargin=10pt,
  innerbottommargin=5pt,
  innerleftmargin=5pt,
  innerrightmargin=5pt,
]{lemmaframe}

\newmdenv[
  backgroundcolor=magenta!10,
  linecolor=magenta!100,
  linewidth=0.8pt,
  skipabove=2pt,
  skipbelow=2pt,
  innertopmargin=10pt,
  innerbottommargin=5pt,
  innerleftmargin=5pt,
  innerrightmargin=5pt,
]{corollaryframe}

\newmdenv[
  backgroundcolor=lime!10,
  linecolor=lime!100,
  linewidth=0.8pt,
  skipabove=2pt,
  skipbelow=2pt,
  innertopmargin=10pt,
  innerbottommargin=5pt,
  innerleftmargin=5pt,
  innerrightmargin=5pt,
]{exampleframe}

\newmdenv[
  backgroundcolor=pink!10,
  linecolor=pink!100,
  linewidth=0.8pt,
  skipabove=2pt,
  skipbelow=2pt,
  innertopmargin=10pt,
  innerbottommargin=5pt,
  innerleftmargin=5pt,
  innerrightmargin=5pt,
]{notationframe}

\newmdenv[
  backgroundcolor=violet!10,
  linecolor=violet!100,
  linewidth=0.8pt,
  skipabove=2pt,
  skipbelow=2pt,
  innertopmargin=10pt,
  innerbottommargin=5pt,
  innerleftmargin=5pt,
  innerrightmargin=5pt,
]{claimframe}

\newmdenv[
  backgroundcolor=salmon!10,
  linecolor=salmon!100,
  linewidth=0.8pt,
  skipabove=2pt,
  skipbelow=2pt,
  innertopmargin=10pt,
  innerbottommargin=5pt,
  innerleftmargin=5pt,
  innerrightmargin=5pt,
]{problemframe}

\newmdenv[
  backgroundcolor=lavender!10,
  linecolor=lavender!100,
  linewidth=0.8pt,
  skipabove=2pt,
  skipbelow=2pt,
  innertopmargin=10pt,
  innerbottommargin=5pt,
  innerleftmargin=5pt,
  innerrightmargin=5pt,
]{observationframe}

\newcommand{\superdit}{{GMem}\xspace}
\newcommand{\REPA}{{REPA}\xspace}

\newcommand{\SiT}{{SiT}\xspace}

\newcommand{\NFE}{{NFE}\xspace}

\newcommand{\FID}{{FID}\xspace}
\newcommand{\Epoch}{{Epochs}\xspace}

\icmltitlerunning{GMem: A Modular Approach for Ultra-Efficient Generative Models}

\begin{document}

\twocolumn[
\icmltitle{GMem: A Modular Approach for Ultra-Efficient Generative Models}

\icmlsetsymbol{equal}{*}

\begin{icmlauthorlist}
  \icmlauthor{Yi Tang}{equal,sew}
  \icmlauthor{Peng Sun}{equal,sew,zju}
  \icmlauthor{Zhenglin Cheng}{equal,sew,zju}
  \icmlauthor{Tao Lin}{sew,rcif}
\end{icmlauthorlist}

{%
\renewcommand\twocolumn[1][]{#1}%
\begin{center}
  \centering
  \vspace{1em}
\centering
\begin{minipage}{0.37\linewidth}
	\centering
	\includegraphics[width=\linewidth]{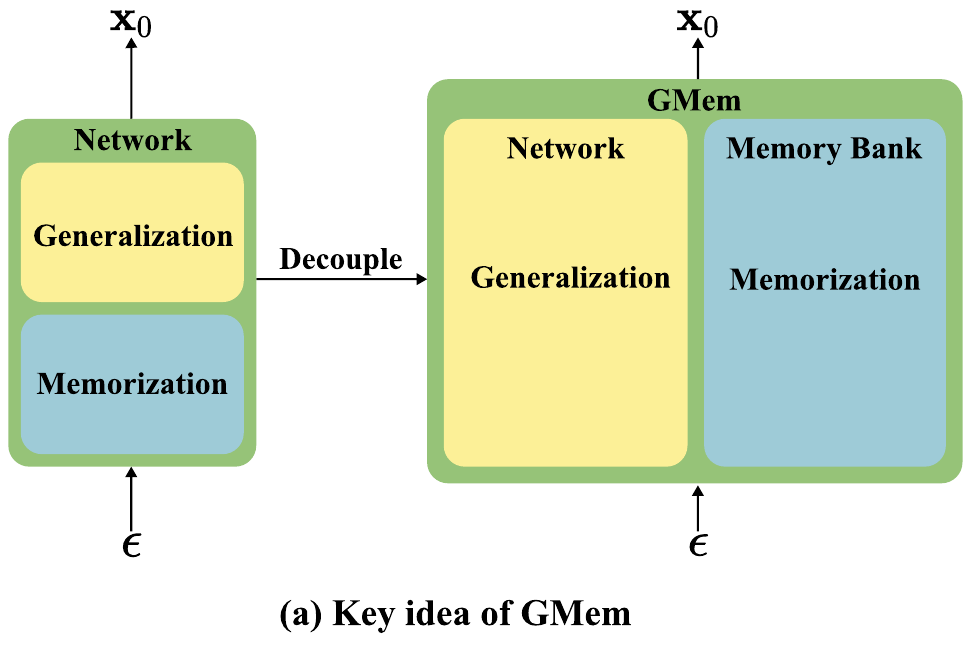}
\end{minipage}
\begin{minipage}{0.3\linewidth}
	\centering
	\includegraphics[width=\linewidth]{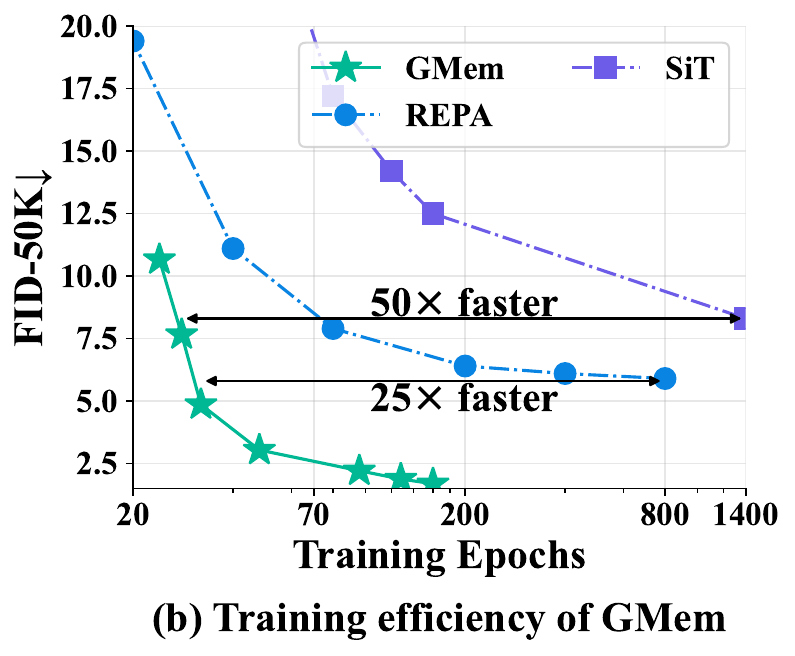} \\
\end{minipage}
\begin{minipage}{0.3\linewidth}
	\centering
	\includegraphics[width=\linewidth]{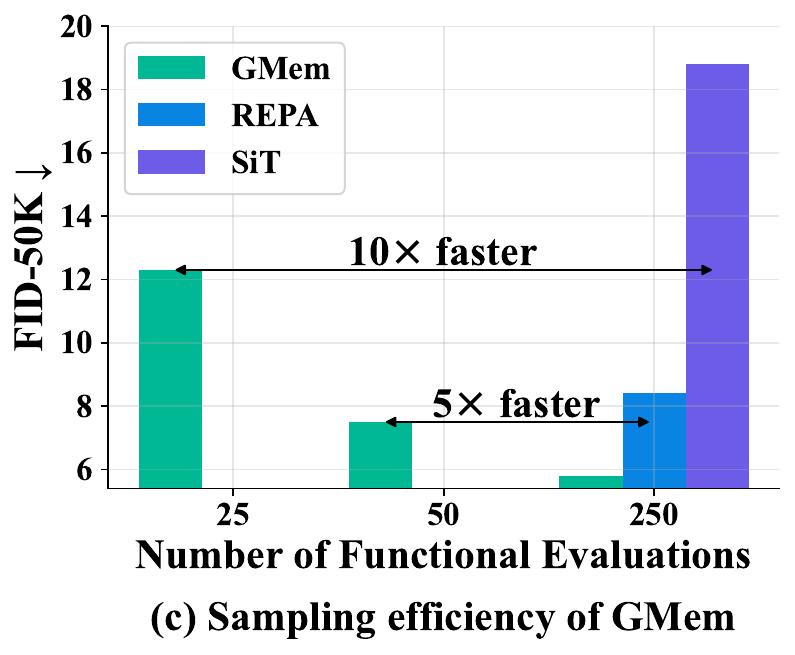} \\
\end{minipage}

\vspace{-1.em}
\captionof{figure}{
	\small
	\textbf{\superdit Significantly enhances training and sampling efficiency of diffusion models on ImageNet \(256\times256\).}
	We propose decoupling memorization capabilities from the model by implementing a separate, immutable memory bank that preserves essential data information.
	\textbf{Sub-figure (a)} highlights the core concept of {\superdit}, where \(\mepsilon\) denotes input noise and \(\xx_0\) represents generated samples. In {\superdit}, we disentangle generalization and memorization capabilities, assigning memorization to an external memory bank \(\mathbf{M}\). This decoupling reduces computational and capacity overhead, thus accelerating the process.
	\textbf{Sub-figure (b)} demonstrates the training efficiency of {\superdit} on ImageNet \(256\times256\). At an \FID$=4.86$, {\superdit} achieves over \(25 \times\) speedup compared to REPA~\citep{yu2024representation}. At an \FID$=7.66$, it achieves over \(50\times\) speedup relative to SiT~\citep{ma2024sit}.
	\textbf{Sub-figure (c)} illustrates sampling efficiency. At the same \FID target, {\superdit} requires \(5\times\) fewer NFEs compared to REPA and \(10\times\) fewer NFEs compared to SiT.
}
\label{fig:motivation}

\end{center}
}

\icmlaffiliation{rcif}{Research Center for Industries of the Future, Westlake University}
\icmlaffiliation{sew}{School of Engineering, Westlake University}
\icmlaffiliation{zju}{Zhejiang University}

\icmlcorrespondingauthor{Tao Lin}{lintao@westlake.edu.cn}

\icmlkeywords{Machine Learning, ICML}

\vskip 0.3in
]

\printAffiliationsAndNotice{\icmlEqualContribution} %

\setlength{\parskip}{2pt plus4pt minus5pt}

\begin{abstract}
  Recent studies indicate that the denoising process in deep generative diffusion models implicitly learns and memorizes semantic information from the data distribution.
  These findings suggest that capturing more complex data distributions requires larger neural networks, leading to a substantial increase in computational demands, which in turn become the primary bottleneck in both training and inference of diffusion models.
  To this end, we introduce {\superdit}: A Modular Approach for Ultra-Efficient Generative Models.
  Our approach {\superdit} decouples the memory capacity from model and implements it as a separate, immutable memory set that preserves the essential semantic information in the data.
  The results are significant: {\superdit} enhances both training, sampling efficiency, and diversity generation.
  This design on one hand reduces the reliance on network for memorize complex data distribution and thus enhancing both training and sampling efficiency.
  On ImageNet at \(256 \times 256\) resolution, {\superdit} achieves a \(50\times\) training speedup compared to SiT, reaching \textbf{\FID$=7.66$} in fewer than $28$ epochs (\textbf{$\sim$$4$ hours} training time)\footnote{All training time measurements are obtained on an $8\times$H800 GPU cluster.}, while SiT requires $1400$ epochs.
  Without classifier-free guidance, {\superdit} achieves state-of-the-art (SoTA) performance \textbf{\FID$=1.53$} in $160$ epochs with \textbf{only $\sim20$ hours} of training, outperforming LightningDiT which requires $800$ epochs and $\sim$$95$ hours to attain \FID$=2.17$.

  Additionally, the memory bank supports training-free adaptation to new images not present in the training set.
  Our code is available at \url{https://github.com/LINs-lab/GMem}.
\end{abstract}

\setlength{\parskip}{2pt plus2pt minus0pt}

\section{Introduction}
\label{sec:intro}

Deep generative models based on denoising~\citep{yang2023diffusion,ho2020denoising,song2020denoising,nichol2021improved,choi2021ilvr,li2024autoregressive,chen2024deconstructing,li2023self}---which prioritize the generation of high-quality, realistic data---have achieved notable success within the deep learning community.
These methods demonstrate exceptional performance in complex tasks such as zero-shot text-to-image and video generation \citep{podell2023sdxl, saharia2022photorealistic, esser2024scaling, moviegen, videoworldsimulators2024}.
However, training and sampling diffusion models suffer from the issue of high computational burden~\citep{karras2022elucidating,karras2024analyzing}, due to the necessity of using neural networks with larger capacities for better empirical performance~\citep{karras2022elucidating,karras2024analyzing}.

In the meanwhile, a line of research examines the diffusion models from the perspective of representation learning~\citep{li2023your, xiang2023denoising, chen2024deconstructing, mukhopadhyay2023diffusion}, where diffusion models can capture semantic features within their hidden layers, with advanced models producing even stronger representations \citep{xiang2023denoising}.
\citet{yang2023diffusion} illustrate that diffusion models strike a balance between learning meaningful features and regularizing the model capacity.
\citet{kadkhodaie2023generalization} further interpret the generalization behavior in diffusion models, indicating that diffusion models aim to learn and encapsulate semantic information from the training data, subsequently generalizing to the true data distribution.
\looseness=-1

In this paper, we build upon the insights from prior works~\cite{yang2023diffusion, kadkhodaie2023generalization} and propose a novel conjecture: \emph{the functionality of diffusion models can be implicitly decomposed into two distinct components: (i) memorization of semantic information and (ii) generalization to the true data distribution.}
We further posit that the intrinsic generation process of diffusion models can be interpreted as a transformation of noisy input samples into the model's internal semantic representation, which is subsequently ``decoded'' into the target data distribution.
To elaborate,
\looseness=-1
\begin{enumerate}[label=(\alph*), nosep, leftmargin=16pt]
  \item During the training stage, diffusion models rely on substantial neural network capacity and computational resources to extract and memorize semantic information from noisy samples;
        \looseness=-1
  \item
        During the sampling/inference stage, the models typically remap Gaussian noise to the internal semantic distribution before reconstructing the data distribution, rather than directly transforming noise into data.
        \looseness=-1
\end{enumerate}
Building upon the aforementioned analyses, the efficiency challenges of training and sampling diffusion models stem from the need to learn and memorize internal semantic information.
This process inherently incurs high computational costs and imposes significant demands on model capacity.

As a remedy, we propose \superdit, a new paradigm that decouples the memorization capability from the neural network by storing semantic information in an external memory bank.
This separation eliminates the need for the neural network itself to memorize complex data distributions, enabling significant improvements:
(1) accelerated training and sampling by avoiding the direct fitting of intricate semantic distributions;
(2) flexible adjustment of memorization capacity in a training-free manner;
(3) enhanced generalization, allowing the generation of images beyond the training set;
and (4) reduced storage overhead by offloading memorization to the memory bank.
Together, \superdit provides a scalable and efficient framework for generative tasks.

Extensive experiments across diverse diffusion backbones~\citep{ma2024sit,yao2025reconstruction} and image tokenizers~\citep{yao2025reconstruction,chen2024deep} validate the effectiveness and efficiency of \superdit. On ImageNet \(256 \times 256\), \superdit achieves SoTA performance \FID$=1.53$ in $160$ epochs with \textbf{only $\sim$$20$ hours training time} without classifier-free guidance.

Under the same number of training epochs as \(80\), {\superdit} enhances the sampling efficiency by $10\times$, where {\superdit} requires only \(25\) sampling steps to achieve an \FID$=12.3$ while the vanilla SiT needs \(250\) sampling steps, yielding \textbf{$10 \times$ sampling time savings}.

\begin{figure*}[h]
  \centering
  \includegraphics[width=\linewidth]{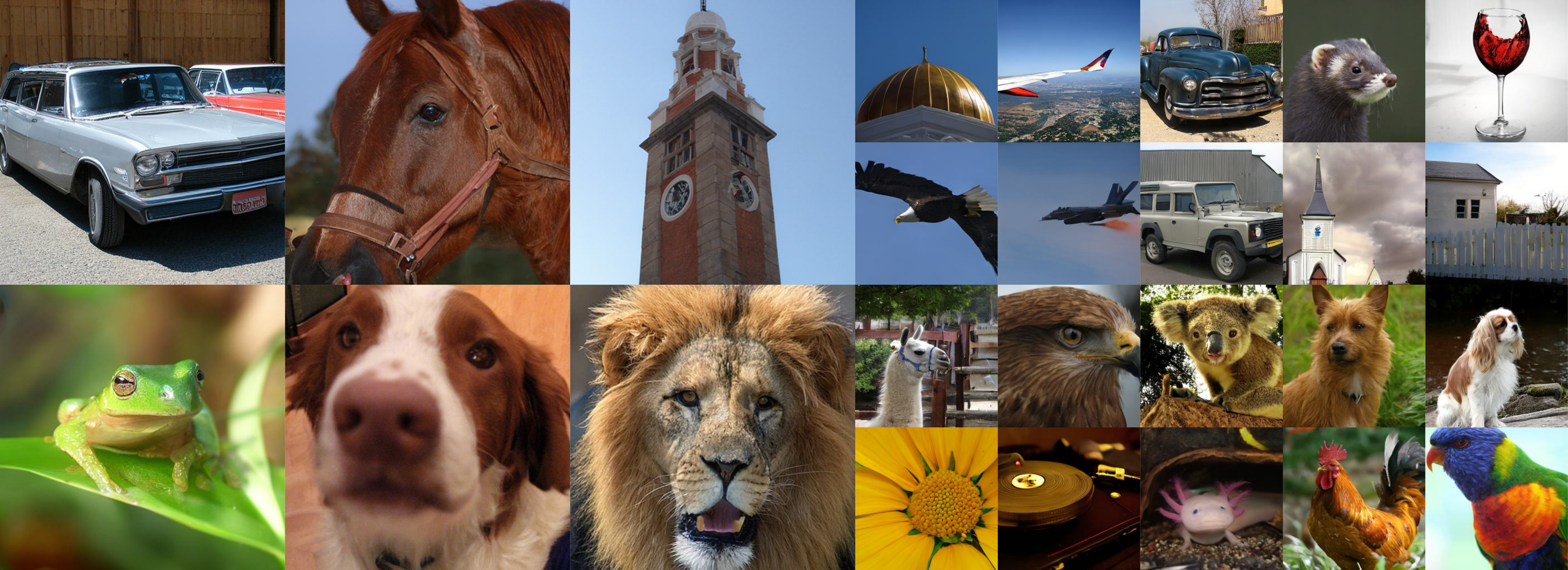}
  \caption{\small
  \textbf{Selected samples on ImageNet \(512 \times 512\) and \(256 \times 256\).}
  This figure presents images generated by \superdit under two experimental settings: (1) For ImageNet $256\times256$, \superdit was trained for $160$ epochs and sampled via Euler method (\NFE$=100$), achieving an \FID$=1.53$ without classifier-free guidance. (2) For ImageNet $512\times512$, training extended to $400$ epochs with identical sampling settings, yielding \FID$=1.89$.
  }

  \label{fig:selected_figs}
  \vspace{-10px}
\end{figure*}

We outline the main contributions of this paper below:
\begin{enumerate}[label=(\alph*), nosep, leftmargin=16pt]
  \item
        We propose a novel conjecture that interprets the functionality of diffusion models into two distinct components: (i) memorization of semantic information and (ii) generalization to the true data distribution.
        \looseness=-1
  \item We introduce {\superdit}, a simple yet highly effective approach that decouples the memorization capability of diffusion models into an external memory bank.
        This separation alleviates the memorization burden on the neural networks and enhances generalization flexibility.
        \looseness=-1
  \item {\superdit} significantly enhances efficiency and quality of the SoTA diffusion models, for both training and inference (see \figref{fig:motivation} and \figref{fig:selected_figs}).
\end{enumerate}

\section{Related work}
\label{sec:relatedwork}

\paragraph{Generative models.}
Generative models--—including Generative Adversarial Networks (GANs)~\citep{goodfellow2014generative,sauer2022stylegan-xl,xiao2021tackling}, Variational Autoencoders (VAEs)~\citep{kingma2013auto, he2022masked}, flow-based methods, and diffusion-based methods~\citep{ho2020denoising,dhariwal2021diffusion,mittal23Diffusion}—--aim to learn the data distribution $p(\xx)$ and generate data through sampling, achieving remarkable performance in producing realistic samples~\citep{li2312return}.
Recently, diffusion-based methods employ stochastic interpolation to model a forward process and then reverse the Gaussian distribution back to the original image space, generating realistic samples.
These methods achieve SoTA results in deep generative modeling and are the focus of this study~\citep{mittal23Diffusion, song2020improved, durkan2021maximum}.

Diffusion models face computational challenges due to high training cost and instability~\citep{yu2024representation,song2020improved} and high sampling costs from multi-step generation~\citep{lu2024simplifying}, driving extensive research to accelerate both processes.
For example, REPA~\citep{yu2024representation} leverages external visual representations to speed up training.
LightningDiT~\citep{yao2025reconstruction} accelerates training by aligning the latent space of vision tokenizer (i.e. VA-VAE) with pretrained vision encoder.
\superdit instead introduces a novel decomposition of diffusion models into generalization and memorization components, enabling significantly faster training.

\paragraph{Diffusion modeling and representation learning.}
To overcome the instability and computational inefficiency of diffusion models, recent studies~\citep{yu2024representation, fuest2024diffusion, mittal23Diffusion} start to leverage representation learning to enhance diffusion models.
On the one hand, diffusion models are capable of learning high-quality representations~\citep{yu2024representation}.
For instance,~\citet{tang2023emergent} demonstrate that feature maps extracted from diffusion networks can establish correspondences between real images, indicating a strong alignment between the learned representations with actual image.
Furthermore, \citet{yang2023diffusion} conduct a detailed analysis of the trade-off between the quality of learned representations and the penalization of the optimal parameter spectrum.

On the other hand, well-trained representation models can improve performance and expedite the training of diffusion models.
\citet{mittal23Diffusion} accomplish this by adjusting the weighting function in the denoising score matching objective to enhance representation learning.
REPA~\citep{yu2024representation} introduces an alignment loss for intermediate layer representations, significantly accelerating the training process by over $17.5$ times.

\paragraph{External representation-augmented diffusion models via retrieval and generation.}
Retrieval-Augmented Generation (RAG) enhances generation quality by integrating external knowledge.
IC-GAN~\citep{casanova2021icgan} augments image generation by conditioning on neighborhood instances retrieved from the training dataset.
However, using only training images limits generalization.
To address this issue, KNN-Diffusion~\citep{sheynin2022knn} and RDM~\citep{blattmann2022rdm} employ large external memory sets, guiding generation via kNN retrieval during training and inference.
Similarly, \citet{chen2022reimagen} and \citet{li2022memory} leverage a set of text-image pairs with cross-modality retrieval, improving generation performance on rare images.

Despite their advantages, RAG methods encounter two key challenges: (1) substantial storage demands for large memory sets, and (2) increased computational costs during retrieval.
We further contend that over-reliance on training sets restricts generalization capabilities~\citep{blattmann2022rdm}.
By employing a masking strategy (see our {\secref{sec:ablation_masking}}), we mitigate this dependency without incurring additional computational or storage overhead, thereby improving both generalization and training/inference efficiency.
\looseness=-1

Representations for generation augmentation can also be obtained from representation generators.
For instance, RCG~\citep{li2023rcg} employs a representation generator to produce 1D ``memory snippets'' to guide diffusion models.
Although RCG reduces the need to store large-scale memory sets, it encounters two primary challenges:
(1) the requirement for additional training and sampling processes for the representation generator, increasing computational demands;
(2) the necessity to retrain the representation generator to incorporate knowledge of new classes or artistic style transfers, thereby limiting its generalization capability.
To overcome these limitations, \secref{sec:exp_diversity} presents an efficient, training-free method for incorporating additional knowledge into the memory bank.

\section{Preliminary}
\label{sec:preliminary}
In this section, we provide a concise introduction to the training and sampling processes of flow-based and diffusion-based models.
See \appref{deriveation} for more details.

Both diffusion-based~\citep{ho2020denoising,dhariwal2021diffusion} and flow-based models~\citep{ma2024sit} derive their training procedures from a deterministic $ T $-step noising process applied to the original data~\citep{ma2024sit}, formalized as: \looseness=-1
\begin{align}
  \label{eq:diffusion_forward}
  \xx_t = \alpha_t \xx_0 + \sigma_t \mepsilon, \quad \mepsilon \sim \mathcal{N}(0, \mathbf{I}) \,,
\end{align}
where \( \xx_t \) represents the noisy data at time \( t \), \( \xx_0 \sim p(\xx) \) is a real data sample from the true distribution, \( \alpha_t \) and \( \sigma_t \) are time-dependent decreasing and increasing functions respectively satisfying \( \alpha_t^2 + \sigma_t^2 = 1 \).

According to \eqref{eq:diffusion_forward}, each marginal probability density \( p_t(\xx_t) \) corresponds to the distribution of a Probability Flow Ordinary Differential Equation~\citep{song2020score} (PF ODE) with velocity field \( \vv(\xx, t) \), defined as:
\begin{align}
  \label{eq:ODE}
  \vv(\xx, t) = \dot{\alpha}_t \, \mathbb{E}[ \xx_0 \mid \xx_t = \xx ] - \dot{\sigma}_t \, \mathbb{E}[\mepsilon \mid \xx_t = \xx ] \,,
\end{align}
where \( \dot{\alpha}_t = \frac{\mathrm{d} \alpha_t}{\mathrm{d} t} \) and \( \dot{\sigma}_t = \frac{\mathrm{d} \sigma_t}{\mathrm{d} t} \).
By solving this ODE starting from \( \xx_T = \mepsilon \sim \mathcal{N}(0, \mathbf{I}) \), we obtain the probability density function \( p_0(\xx_0) \), which can be used to estimate the ground-truth data distribution \( p(\xx) \).

Alternatively, the aforementioned noise-adding process can be formalized as a Stochastic Differential Equation~\citep{song2020score, song2020denoising} (SDE):
\begin{align}
  \label{eq:SDE}
  \mathrm{d} \xx_t = \boldsymbol{m}(\xx_t, t) \, \mathrm{d} t + g(t) \, \mathrm{d} \mathbf{W}_t \,,
\end{align}
where \( \mathbf{W}_t \) is a Wiener process~\citep{hitsuda1968representation}, \( \boldsymbol{m}(\xx_t, t) \) is the drift coefficient defined as \( \boldsymbol{m}(\xx_t, t) = -\frac{1}{2} \beta(t) \xx_t \), and \( g(t) \) is the diffusion coefficient, set as \( g(t) = \sqrt{\beta(t)} \) with \( \beta(t) \) being a time-dependent positive function controlling the noise schedule.

The corresponding reverse process is represented by the reverse-time SDE:
\begin{align}
  \label{eq:reverse_SDE}
  \mathrm{d} \xx_t = \left[ \boldsymbol{m}(\xx_t, t) - g(t)^2 s(\xx_t, t) \right] \mathrm{d} t + g(t) \, \mathrm{d} \bar{\mathbf{W}}_t \,,
\end{align}
where \( \bar{\mathbf{W}}_t \) is a reverse-time Wiener process, and \( s(\xx_t, t) \) is the score function, defined by the gradient of the log probability density:
\begin{align}
  \label{eq:score_standard}
  s(\xx_t, t) = \nabla_{\xx_t} \log p_t(\xx_t) = - \frac{1}{\sigma_t} \mathbb{E}\left[ \mepsilon \mid \xx_t = \xx \right] \,.
\end{align}

By solving the reverse-time SDE in~\eqref{eq:reverse_SDE}, starting from the initial state \( \xx_T = \mepsilon \sim \mathcal{N}(0, \mathbf{I}) \), we can obtain \( p_0(\xx_0) \), thereby estimating the true data distribution \( p(\xx) \).

\section{Methodology}
\label{sec:methodology}

\begin{algorithm}[t!]
    \caption{\small
        Training {\superdit} using memory bank $\mathbf{M}$
    }
    \label{alg:training}
    \begin{algorithmic}
        \Procedure{Train {\superdit}}{$\vv_{\mtheta}, \mathcal{D}, \mathbf{M}, T, \alpha_t, \sigma_t$}
        \State \textbf{Initialize} model parameters $\mtheta$
        \For{each training iteration}
        \State \textbf{Sample} a batch of data $\xx_0 \sim \mathcal{D}$
        \State \textbf{Sample} timesteps $t \sim \{0, \dots, T\}$ uniformly
        \State \textbf{Generate} noise $\mepsilon \sim \mathcal{N}(0, \mathbf{I})$
        \State \textbf{Compute} noisy data $\xx_t = \alpha_t \xx_0 + \sigma_t \mepsilon$
        \State \textbf{Sample} memory snippet $\mathbf{s} \sim \mathbf{M}$
        \State \textbf{Predict} velocity $\vv_{\mtheta}(\xx_t, t, \mathbf{s})$
        \State \textbf{Compute} loss using~\eqref{eq:modified_df_loss}
        \State \textbf{Backpropagate} and update $\mtheta$ using Optimizer
        \EndFor
        \EndProcedure
    \end{algorithmic}
\end{algorithm}

{\superdit} is motivated by the observation that diffusion models inherently encode two distinct capabilities: \emph{generalization} and \emph{memorization} of semantic information within data~\citep{yu2024representation,yang2023diffusion,xiang2023denoising,chen2024deconstructing}.
We introduce a novel paradigm that decouples these functions: neural networks handle generalization, while an external memory bank stores semantic information.
\looseness=-1

Current generative models~\citep{ho2020denoising,song2020improved} typically rely on one neural network to simultaneously achieve both generalization and memorization of data distributions.
However, the capacity-constrained modern diffusion models, such as UNet and Transformer-based networks~\citep{ho2020denoising,peebles2023scalable}, face two critical limitations~\citep{kadkhodaie2023generalization}: (1) insufficient model parameters to memorize complex data distributions, and (2) computationally expensive parameter optimization for memorization.

\begin{figure}[t]
  \centering
  \includegraphics[width=\linewidth]{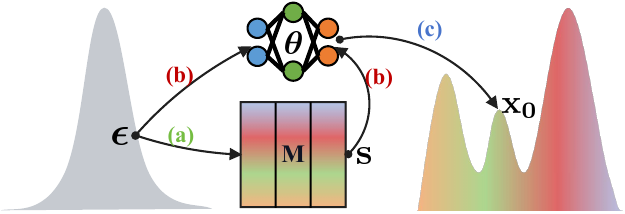}
  \caption{\small
    \textbf{Data generation via \superdit-enhanced diffusion models.}
    \textcolor{green}{(a)} Sampled noise $\mepsilon$ is used to index a memory snippet from the memory bank. \textcolor{red}{(b)} Both the sampled noise $\mepsilon$ and the memory snippet $\mathbf{s}$ are simultaneously fed into the neural network. \textcolor{blue}{(c)} The neural network generates data using SDE or ODE solvers.
  }
  \label{fig:framework}
\end{figure}

To overcome these limitations, we propose offloading the memorization function to an external memory bank, thereby enhancing training efficiency and distribution capture capabilities.
The proposed framework is illustrated in \figref{fig:framework}.
\looseness=-1

\subsection{External Memory Bank}
Building on~\citet{kadkhodaie2023generalization}'s insight that diffusion models achieve generalization through geometry-adaptive harmonic representations, we design a memory bank for diffusion models to:
(a) supply essential semantic information for generating high-quality, realistic, and distribution-aligned images,
and (b) exclude excessive details to prevent overfitting to the training data while maintaining robust generalization.
\looseness=-1

We formalize the memory bank as a matrix \(\mathbf{M} \in \mathbb{R}^{n \times m}\), consisting of \(n\) unit-norm memory snippets: \looseness=-1
\begin{equation}
  \mM = \begin{bmatrix}
    \mathbf{s}_1, \mathbf{s}_2, \ldots, \mathbf{s}_n
  \end{bmatrix}^\top \,, \quad |\mathbf{s}_i|=1 \,.
\end{equation}
We employ a representation model \(\mf\) such that for any input \(\mathbf{x} \sim D \), the normalized feature \(\nicefrac{\mf(\mathbf{x})}{\|\mf(\mathbf{x})\|}\) corresponds to a vector \( \mathbf{s} \in \mathbf{M} \).
Through optimization, we ensure that \(\mathbf{M}\) fully captures the semantic information of $D$.
\looseness=-1

\paragraph{Representation models for memory bank construction.}
{\superdit} supports diverse representation models $\mf$, each varying in their ability to capture image information, which also directly influences generative model performance.
We employ self-supervised representation models for two key reasons:
\begin{enumerate}[label=(\alph*), nosep, leftmargin=16pt]
  \item \citet{yu2024representation} demonstrate that self-supervised models can expedite the training of diffusion models.
        It motivates to use it in {\superdit}.
  \item Self-supervised models capture semantic information more effectively than supervised alternatives~\citep{bordes2022high,zimmermann2021contrastive, sun2024efficiency}.
\end{enumerate}
We examine multiple models, including CLIP visual encoder \citep{radford2021clip}, to construct the memory bank (see \appref{app:text2image}).

\paragraph{Reducing network dependency on memory bank.}
Directly using memory snippets to guide network generation can lead to overfitting, as the network may rely too heavily on the strong guidance provided, impairing its ability to generalize to unseen samples~\citep{blattmann2022rdm}.
To address this issue, we propose a feature-level random masking strategy. Instead of utilizing the complete memory snippets during training, we randomly mask a subset of dimensions by setting them to zero.
For instance, given a memory snippet with 512 dimensions, we can randomly select 256 dimensions to mask.

We will detail the performance of various masking strategies and ratios in \secref{sec:ablation_masking} and discuss how to select the most appropriate masking strategy.

\subsection{Training Objective}
Memory bank provides semantic information about the data distribution, aiding both training and inference phases in diffusion models.
To integrate this, we adapt the training loss of diffusion models as follows:
\begin{equation}{
    \label{eq:modified_df_loss}
    \cL(\mtheta) = \int_{0}^{T} \E{
      \norm{\vv_{\mtheta}(\xx_t,\mathbf{s},t) - \dot{\alpha}_t \xx_0 - \dot{\sigma}_t \mepsilon}^2 dt
    } \,,
  }
\end{equation}
where $\xx_0 \sim D, \mepsilon \sim \cN (0, \mI) ,  \dot{\alpha}_t = \frac{\mathrm{d} \alpha_t}{\mathrm{d} t} ,  \dot{\sigma}_t = \frac{\mathrm{d} \sigma_t}{\mathrm{d} t}$ and $\vv_{\mtheta}$ is the velocity estimated by SiT backbone.
For fair comparison, we retain the same $\alpha_t$ and $\sigma_t$ settings as in previous work \citep{ma2024sit, yu2024representation}.

\subsection{Sampling with Memory Bank}
The sampling stage of diffusion models~\citep{ho2020denoising,lu2024simplifying} transforms Gaussian noise into the true data distribution through an end-to-end framework, utilizing SDE or ODE solvers (see \secref{sec:preliminary}) to iteratively refine the noise.
Similarly, \superdit, equipped with a memory bank, follows this process by feeding noise and corresponding memory snippet~\(\mathbf{s}\) into the neural network.

However, akin to~\citet{blattmann2022rdm,casanova2021icgan}, the memory bank requires explicitly storage and retrieval during inference.
For large-scale datasets like ImageNet, this results in prohibitive storage demands and retrieval costs.
Additionally, memory snippets often exhibit significant redundancy (e.g., snippets from the same class are highly correlated), leading to unnecessary storage when all features are retained directly.

To reduce redundancy in storing memory snippets and minimize storage, we propose an efficient storage strategy.
\paragraph{Decomposing the memory bank into two compact matrices.}
\label{sec:dmd}
Let \(\mathbf{M} \in \mathbb{R}^{N \times d}\) represent \(N\) memory snippets \(\ss\), each of dimension \(d\).
We first compute the mean vector \(\mmu \in \mathbb{R}^d\) and centralize \(\mathbf{M}\) as \(\mathbf{M}_{\text{centered}} = \mathbf{M} - \mmu\).
Applying SVD~\citep{eckart1936approximation}
to it yields $\mathbf{M}_{\text{centered}} = \mathbf{U} \mathbf{\Sigma} \mathbf{V}^\top$, where \(\mathbf{U} \in \mathbb{R}^{N \times r}\), \(\mathbf{\Sigma} \in \mathbb{R}^{r \times r}\), and \(\mathbf{V} \in \mathbb{R}^{d \times r}\), with \(r < \min(N,d)\) as the target latent dimension.
\looseness=-1

By distributing the singular values, we form a coefficient matrix \(\mathbf{C} = \mathbf{U}\,\mathbf{\Sigma}^{1/2} \in \mathbb{R}^{N \times r} \) and a basis matrix \(\mathbf{B} = \mathbf{V}\,\mathbf{\Sigma}^{1/2} \in \mathbb{R}^{d \times r} \).
This results in a low-rank approximation:
\begin{equation}
  \mathbf{M} \gets \mathbf{C}\,\mathbf{B}^\top \,,
\end{equation}
reducing storage complexity from \(\mathcal{O}(N d)\) to \(\mathcal{O}(N r + d r)\).

The matrix \(\mathbf{B}\) acts as a compact, fixed basis encoding global structure, whereas \(\mathbf{C}\) flexibly stores snippet-specific coefficients.
During inference, retrieving a snippet involves looking up its coefficients in \(\mathbf{C}\) and transforming them via \(\mathbf{B}\).
Specifically, the $i$-th memory snippet $\ss_i$ can be reconstructed as:
\begin{equation}
  \ss_i = \mathbf{c}_i \mathbf{B}^\top + \mmu\,,
\end{equation}
where \(\mathbf{c}_i \in \mathbb{R}^{1 \times r}\) is the coefficient vector corresponding to the $i$-th snippet from coefficient matrix $\mC$.

\subsection{Modular Memory Bank Enables Generalization}
The core insight of \superdit lies in the decoupling of memorization and generalization in a generative model, enabling generalization beyond training data.
This is achieved through two approaches: (I) incorporating external knowledge by introducing novel images absent from the original dataset, and (II) manipulating internal knowledge by combining existing memory snippets into new compositions.
While our memory bank provides compact storage and flexibility for new snippets integration, we note that advancing the modularity of the memory bank remains future work.

\paragraph{Aspect I: external knowledge manipulation.}
To incorporate external knowledge outside the training dataset, we project the feature vector \(\mf(\xx_{\text{new}}) \in \mathbb{R}^d\) onto the existing coefficient matrix \(\mC\).
We calculate the coefficients \(\mathbf{c}_{\text{new}} \in \mathbb{R}^r\) of the centered feature vector by projecting it onto the basis matrix \(\mathbf{B}\):
\begin{equation}
  \mathbf{c}_{\text{new}} = \nicefrac{(\mf(\xx_{\text{new}}) -\mmu)\,\mathbf{B}}{\mathbf{S}} \,,
\end{equation}
where \(\mathbf{S}\) (the diagonal of \(\mathbf{\Sigma}\)) stores singular values\footnote{These singular values are several floating-point numbers with negligible storage cost.}.
By appending \(\mathbf{c}_{\text{new}}\) to \(\mathbf{C}\)
, we expanding the memory bank with negligible overhead.
This process seamlessly integrates new snippets not present in the training dataset, allowing the model to utilize the network's generalization capabilities for generating new samples without additional training.

\paragraph{Aspect II: internal knowledge manipulation.}
We generate new memory snippets by interpolating between existing ones.
Given two snippets indexed by \(i\) and \(j\), we construct a new coefficient vector:
\begin{equation}
  \mathbf{c}_{\text{new}} = \alpha\,\mathbf{c}_i + (1 - \alpha)\,\mathbf{c}_j, \quad \alpha \in [0, 1] \,.
\end{equation}
Appending \(\mathbf{c}_{\text{new}}\) to \(\mathbf{C}\) yields a latent interpolation in \(\mM\) without modifying \(\mathbf{B}\).
This approach enables training-free style transfer and compositional generalization by exploring linear paths between coefficients of different memory snippets, effectively creating novel samples from the internal knowledge encoded in \(\mathbf{C}\).

In \secref{sec:exp_diversity}, we demonstrate the effectiveness of both exact projection and interpolation for expanding the memory bank, enhancing the diversity and flexibility of generated results.
\looseness=-1

\section{Experiments}
\label{sec:experiments}

\subsection{Experimental Setting}
We list the setup below (see more details in \appref{app:experiments}).
\looseness=-1

\paragraph{Datasets.}

For pixel-level generation, we test {\superdit} on CIFAR-10~\citep{krizhevsky2009cifar} due to its diverse classes and its popularity in benchmarking image generation.
We then evaluate {\superdit} on ImageNet $256 \times 256$~\citep{deng2009imagenet} to examine how it models latent space distributions, which is a key focus in recent image generation research~\citep{karras2022elucidating,yu2024representation,ma2024sit,peebles2023scalable}.
Finally, we assess the scalability of {\superdit} to larger resolutions by conducting experiments on ImageNet $512 \times 512$~\citep{deng2009imagenet}.
\looseness=-1

\paragraph{Backbones and visual tokenizers.}
Following prior image generation approaches \citep{yu2024representation}, we primarily use LightningDiT~\citep{yao2025reconstruction} and SiT~\citep{ma2024sit} as the backbone.
We also evaluate the effectiveness of {\superdit} on different visual tokenizer such as SD-VAE~\citep{rombach2022vae}, DC-AE~\citep{chen2024deep}, and VA-VAE~\citep{yao2025reconstruction} in~\tabref{tab:architectures}.

These architectures represent the latest advancements in diffusion and flow-based transformers~\citep{yu2024representation} and are widely adopted in image generation tasks~\citep{han2023svdiff, chen2024pixart, yu2024representation}.

\paragraph{Baselines.}
For a fair comparison, we compare to the SoTA image generation methods on both training efficiency and performance.
Specifically, for pixel-space image generation, we consider the following three categories of baselines:
First, we compare \superdit with traditional generative models, including Diffusion GAN~\citep{xiao2021tackling}, Diffusion StyleGAN~\citep{wang2022diffusion}, DMD2~\citep{yin2024improved}.
We also compare the SoTA diffusion models with UNets, including DDPM~\citep{ho2020denoising}, Score SDE~\citep{song2020score}, EDM~\citep{karras2022elucidating},  DPM-Solver~\citep{lu2022dpm}, ADM~\citep{dhariwal2021diffusion},  EDMv2~\citep{karras2024analyzing},  CTM~\citep{kim2023ctm},  SiD~\citep{zhou2024SiD}, EMD~\citep{karras2022elucidating}.
Finally, we also compare to the SoTA flow-based transformer methods, including DiT~\citep{peebles2023scalable}, SiT~\citep{ma2024sit}, and the most recent yet contemporaneous work REPA~\citep{yu2024representation}.
\looseness=-1

\begin{table*}[t]
    \centering
    \caption{\small
        \textbf{Sampling quality on various datasets.}
        We report the performance of {\superdit} on CIFAR-10 (left), ImageNet $256\times256$ (middle) and ImageNet $512\times512$ (right).
            {\superdit} achieves comparable {\FID} with fewer sampling steps across multiple datasets, further highlighting its advantages in accelerating training.
        All the results reported are w/o classifier-free guidance unless otherwise specified.
    }
    \vspace{-0.5em}
    \label{tab:cifar_in64}
    \begin{minipage}[t]{0.33\textwidth}
        \vspace{0pt}
        \resizebox{0.99\textwidth}{!}{%
            \begin{tabular}{lrr}
                \multicolumn{3}{l}{\textbf{CIFAR-10}}                                                                       \\
                \toprule
                \textbf{METHOD}                               & \textbf{Epoch} ($\downarrow$) & \textbf{FID} ($\downarrow$) \\
                \midrule

                \multicolumn{3}{l}{\textbf{Traditional generative models}}                                                  \\
                \midrule
                BigGAN~\citep{brock2018large}                 & -                             & 14.7                        \\  %
                StyleGAN2~\citep{karras2020training}          & 128                           & 8.32                        \\  %
                \midrule

                \multicolumn{3}{l}{\textbf{Diffusion models (UNets)}}                                                       \\
                \midrule
                DDPM~\citep{ho2020denoising}                  & 100                           & 3.17                        \\
                DDIM~\citep{song2020denoising}                & -                             & 4.04                        \\
                Score SDE (deep)~\citep{song2020score}        & 333                           & 2.20                        \\
                EDM~\citep{karras2022elucidating}             & 400                           & 2.01                        \\
                Diffusion Style-GAN~\citep{wang2022diffusion} & -                             & 3.19                        \\
                Diffusion GAN ~\citep{xiao2021tackling}       & 1024                          & 3.75                        \\

                \midrule

                \multicolumn{3}{l}{\textbf{Diffusion (Transformer)}}                                                        \\
                \midrule
                SiT-XL/2~\citep{ma2024sit}                    & 512                           & 6.68                        \\
                ~~+ \REPA~\citep{yu2024representation}        & 200                           & 4.52                        \\
                ~~+ \superdit(ours)                           & 52                            & \textbf{4.08}               \\
                ~~+ \superdit(ours)                           & 200                           & \textbf{1.59}               \\
                ~~+ \superdit (ours)                          & 450                           & \textbf{1.22}               \\
                \bottomrule
            \end{tabular} %
            \vspace{0.5em}
        }
    \end{minipage}
    \hfill
    \begin{minipage}[t]{0.33\textwidth}
        \vspace{0pt}
        \resizebox{0.97\textwidth}{!}{%
            \begin{tabular}{lrr}
                \multicolumn{3}{l}{\textbf{ImageNet 256$\times$256}}                                                       \\
                \toprule
                \textbf{METHOD}                              & \textbf{Epoch} ($\downarrow$) & \textbf{FID} ($\downarrow$) \\
                \midrule
                \multicolumn{3}{l}{\textbf{Traditional generative models}}                                                 \\
                \midrule
                BigGAN~\citep{brock2018large}                & -                             & 6.96                        \\ %
                VQ-GAN~\citep{esser2021taming}               & -                             & 15.78                       \\% from SD-DiT
                \midrule

                \multicolumn{3}{l}{\textbf{Diffusion models (UNets)}}                                                      \\
                \midrule
                {ADM}~\citep{dhariwal2021diffusion}          & 400                           & 10.94                       \\% paper
                \midrule

                \multicolumn{3}{l}{\textbf{Diffusion models (Transformer)}}                                                \\
                \midrule
                MaskGIT~\citep{chang2022maskgit}             & 300                           & 6.18                        \\ %
                MAGVIT-v2~\citep{yu2023language}             & 270                           & 3.65                        \\ %
                SD-DiT~\citep{zhu2024sd}                     & 480                           & 7.21                        \\ %
                {DiT-XL/2}~\citep{peebles2023scalable}       & 1400                          & 9.62                        \\% SD-DiT
                {SiT-XL/2}~\citep{ma2024sit}                 & 1400                          & 8.30                        \\ %
                {~~+ \REPA}~\citep{yu2024representation}     & 782                           & 5.90                        \\ %
                {FasterDiT}~\citep{yao2024fasterdit}         & 400                           & 7.91                        \\ %
                {LightningDiT}~\citep{yao2025reconstruction} & 800                           & 2.17                        \\ %
                {~~+ \superdit} (ours)                       & 160                           & \textbf{1.53}               \\
                \bottomrule
            \end{tabular} %
        }
    \end{minipage}
    \hfill
    \begin{minipage}[t]{0.33\textwidth}
        \vspace{0pt}
        \resizebox{0.97\textwidth}{!}{%
            \begin{tabular}{lrc}
                \multicolumn{3}{l}{\textbf{ImageNet 512$\times$512}}                                                          \\
                \toprule
                \textbf{METHOD}                                 & \textbf{Epoch} ($\downarrow$) & \textbf{FID} ($\downarrow$) \\
                \midrule

                \multicolumn{3}{l}{\textbf{Traditional generative models}}                                                    \\
                \midrule
                StyleGAN-XL~\citep{sauer2022stylegan-xl}        & -                             & 2.41                        \\
                BigGAN~\citep{brock2018large}                   & 472                           & 9.54                        \\
                \midrule
                \multicolumn{3}{l}{\textbf{Diffusion models (UNets)}}                                                         \\
                \midrule
                ADM~\citep{karras2024analyzing}                 & -                             & 23.2                        \\
                EDM2~\citep{karras2024analyzing}                & 734                           & 1.91                        \\

                \midrule

                \multicolumn{3}{l}{\textbf{Diffusion models (Transformer)}}                                                   \\
                \midrule
                MaskGIT~\citep{chang2022maskgit}                & 300                           & 7.32                        \\
                MAGVIT-v2~\citep{yu2023language}                & 270                           & 3.07                        \\
                DiT-XL~\citep{peebles2023scalable}              & 600                           & 12.03                       \\
                SiT-XL~\citep{ma2024sit} (w/ cfg)               & 600                           & 2.62                        \\
                ~~+ \REPA\citep{yu2024representation}  (w/ cfg) & 200                           & 2.08                        \\
                LightningDiT-XL~\citep{yao2025reconstruction}   & -                             & -                           \\
                ~~+ \superdit (ours)                            & 200                           & 1.93                        \\
                ~~+ \superdit (ours)                            & 400                           & \textbf{1.89}               \\
                ~~+ \superdit (ours) (w/ cfg)                   & 400                           & \textbf{1.71}               \\
                \bottomrule
            \end{tabular} %
        }
    \end{minipage}
    \vspace{-1.5em}
\end{table*}

\paragraph{Metrics.}
In line with prior research~\citep{ma2024sit,hoogeboom2023simple}, all generation quality metrics are reported as {FID-$50\text{K}$} (FID)~\citep{heusel2017fid} scores. %
For CIFAR-10 dataset, we use the training set as the reference set, and for ImageNet $64\times64$ and ImageNet $256\times256$, we use the validation set as the reference set.
Following~\citet{yu2024representation}, we also report the {epochs} as a measure of training efficiency.
With similar \FID scores, the model with less {epochs} is considered more efficient\footnote{
  $64$ {epochs} corresponds to a total of $\sim80\text{K}$ training steps with the batch size of $1024$, and baselines of different batch-sizes or training steps can be converted to {epochs} in a similar way.
}.
Moreover, following~\citep{lu2024simplifying}, we evaluate the efficiency of the sampling by comparing the {Number of Function Evaluations} (\NFE) required to achieve a certain \FID score.
Under our setting, a lower \NFE corresponds to reduced computational costs during sampling, thus making the model more efficient.
\looseness=-1

\paragraph{Implementation details.}
Unless otherwise specified, we adhere to the setup details outlined in \citep{ma2024sit} to ensure a fair comparison. For latent-space generation, we utilize ImageNet \(256 \times 256\) and \(512 \times 512\).
We follow the ADM framework \citep{dhariwal2021diffusion} for additional preprocessing strategies.
Following \citet{ma2024sit}, preprocessed images are fed into SD-VAE~\citep{rombach2022vae} and embedded into the latent space $\zz\in\R^{32\times32\times4}$.
Additionally, we add representation alignment loss intrfoduced from REPA~\citep{yu2024representation} to help the model learn more structured representations across all experiments.
For model configurations, unless otherwise specified, we empolys SiT-XL for CIFAR-10 and LightningDiT-XL for ImageNet $256\times256$ and $512\times512$.

We consistently use the batch size of $128$ for CIFAR-10 following~\citep{song2020score} and $1024$ for ImageNet following~\citep{karras2024analyzing} during training.
Additional experimental implementation details are provided in \appref{app:impl_detail}.

\paragraph{Sampler configurations.} Following SiT~\citep{ma2024sit}, we use SDE solver and set the NFE as $50$ for CIFAR-10 and $100$ for ImageNet $256\times256$ and $512\times512$ by default. More details are provided in \appref{app:impl_detail}.

\begin{figure*}[!t]
  \centering
  \includegraphics[width=\linewidth]{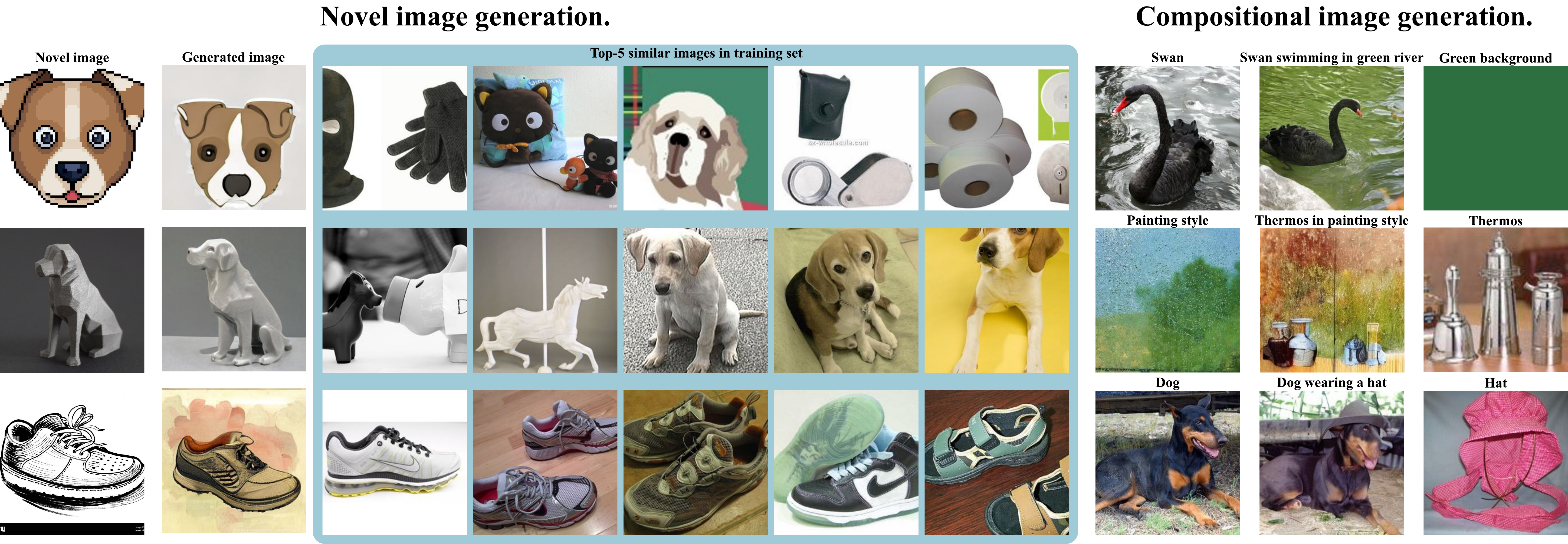}
  \vspace{-2em}
  \caption{\small
    \textbf{Demonstration of novel and compositional image generation via memory bank manipulation.}
    Selected samples from ImageNet \(256\times256\) generated by the \superdit. In the ``Novel image generation'' part, we show the reference image used to build a new memory snippet (left), followed by the generated samples and 5 of the nearest training images, illustrating \superdit's adaptation to external knowledge.
    In the ``Compositional image generation'' examples, two reference images (left and right) form an interpolated image (center), demonstrating \superdit can manipulate internal knowledge to create new concepts.
  }
  \label{fig:novel}
  \vspace{-0.5em}
\end{figure*}

\begin{figure*}[!t]
  \centering
  \includegraphics[width=\linewidth]{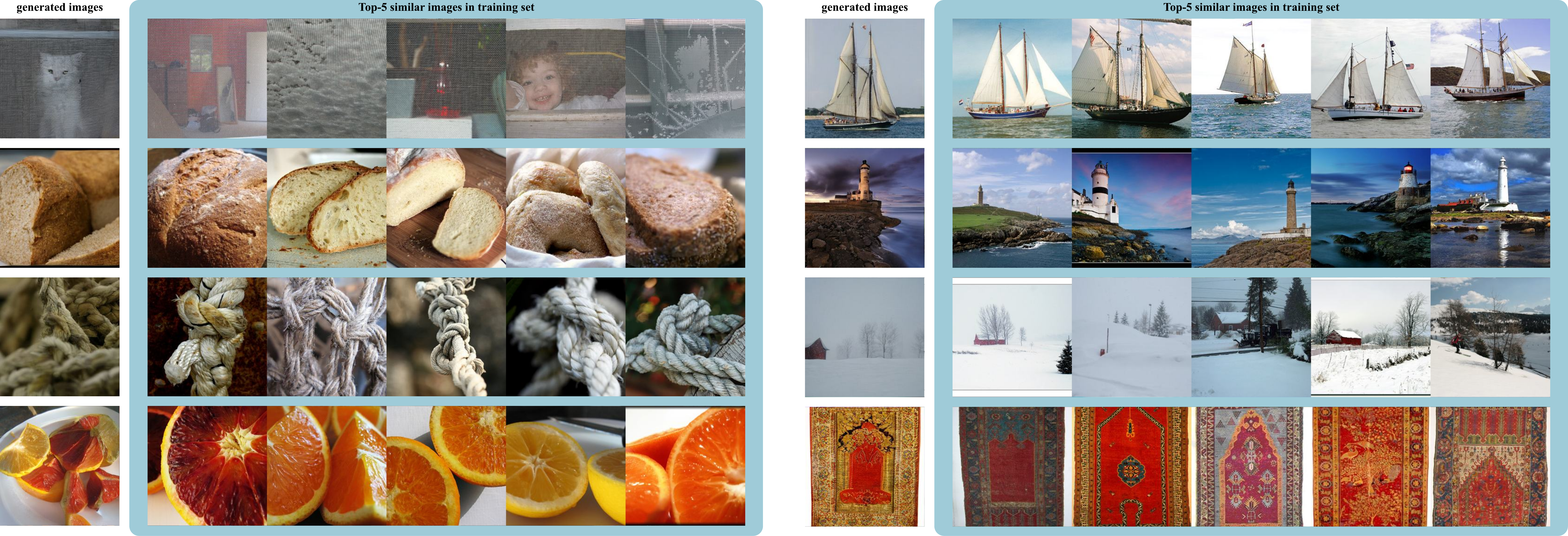}
  \vspace{-2em}
  \caption{\small
    \textbf{Demonstration of diverse generation by {\superdit}.}
    Selected samples from ImageNet \(256\times256\) generated by the {\superdit}. This figure demonstrates the diversity of images generated by {\superdit}, which differ from the original training set in form, style, and color.
    This shows that {\superdit} does not simply memorize images from the training set, but rather generates novel variations.
  }
  \label{fig:diverse}
  \vspace{-0.5em}
\end{figure*}

\subsection{Training and sampling efficiency of {\superdit}}

\begin{table}[t!]
    \centering\tiny
    \captionof{table}{\small
    \textbf{{\superdit} offers 50$\times$ training speedup compared to SiT.}
    This table compares training speedup provided by \superdit to other baselines on ImageNet $256\times256$.
    $\downarrow$ indicates lower is better, and all results are without classifier-free guidance.
    }
    \vspace{-0.10in}
    \resizebox{0.4\textwidth}{!}{%
        \begin{tabular}{lcccr}
            \toprule[0.4pt]
            \textbf{Method} & \textbf{\#Params} & \textbf{Epoch} & \textbf{FID$\downarrow$} \\
            \midrule[0.2pt]
            DiT-XL          & 675M              & {1400}         & {9.62}                   \\
            SiT-XL          & 675M              & {1400}         & {8.61}                   \\
            ~~+ \REPA       & 675M              & {800}          & {5.90}                   \\
            LightningDiT-XL & 675M              & {800}          & {2.17}                   \\
            {~~+ \superdit} & 675M              & {28}           & {7.66}                   \\
            {~~+ \superdit} & 675M              & {32}           & {4.86}                   \\
            {~~+ \superdit} & 675M              & {160}          & {1.53}                   \\
            \bottomrule[0.4pt]
        \end{tabular}
        \label{tab:efficiency}
    }

    \vspace{-0.11in}
\end{table}

\paragraph{{\superdit} is a $\mathbf{50\times}$ training accelerator.}
We evaluate training efficiency and performance across pixel-space and latent-space generation tasks.
As shown in \tabref{tab:cifar_in64} and \tabref{tab:efficiency}, {\superdit} achieves competitive {\FID} scores with significantly fewer training epochs.

\emph{\textbf{Pixel-space generation:}}
On CIFAR-10 in \tabref{tab:cifar_in64}, {\superdit} matches {\REPA}'s performance in just $450$ {epochs}, delivering a $3.85\times$ speedup over {\REPA} and over $10\times$ compared to traditional {\SiT}.
\looseness=-1

\emph{\textbf{Latent-space generation:}}
On ImageNet in \tabref{tab:cifar_in64}, {\superdit} achieves \FID$=1.53$ on ImageNet $256\times 256$ and $1.71$ on ImageNet $512\times 512$ in just $160$ and $400$ epochs, without using classifier-free guidance.
As shown in \tabref{tab:efficiency}, {\superdit} achieves competitive {\FID} scores with significantly fewer epochs than the most efficient baselines {\REPA} and {\SiT}:
(1) With only $32$ epochs (\textbf{$\mathbf{25\times}$ speedup}), {\superdit} achieves \FID$=4.86$, outperforming {\REPA}, which requires $800+$ epochs.
(2) With only $28$ epochs (\textbf{$\mathbf{50\times}$ speedup}), {\superdit} reaches \FID$=7.66$, surpassing {\SiT}'s \FID$=8.61$ at $1{,}400$ epochs.
\looseness=-1

\paragraph{{\superdit} is a $\mathbf{10\times}$ sampling accelerator.}
\begin{figure}[t!]
    \centering\tiny
    \captionof{table}{\small
        \textbf{{\superdit} accelerates sampling by 10$\times$.}
        This table examines the impact of different $\NFE$ choices on $\FID$.
        For a fair comparison with {\REPA}, we use SiT-L/2 as backbone and trained {\superdit} for $\Epoch=20$ (400K iterations with a batch size of 256) without using classifier-free guidance.
        $\downarrow$ means lower is better.
    }
    \vspace{-0.10in}
    \resizebox{0.325\textwidth}{!}{%
        \begin{tabular}{lccr}
            \toprule[0.4pt]
            \textbf{Model} & \textbf{Epoch} & \textbf{NFE$\downarrow$} & \textbf{FID$\downarrow$} \\
            \midrule[0.2pt]
            SiT-L/2        & 20             & 250                      & 18.8                     \\
            ~~+ \REPA      & 20             & 250                      & \pz8.4                   \\
            ~~+ \superdit  & 20             & 250                      & \textbf{5.8}             \\
            ~~+ \superdit  & 20             & 50                       & \textbf{7.5}             \\
            ~~+ \superdit  & 20             & 25                       & \textbf{12.3}            \\
            \arrayrulecolor{black}\bottomrule[0.4pt]
        \end{tabular}
        \label{tab:nfe}
    }

    \vspace{-0.11in}
\end{figure}

By generating competitive samples with significantly fewer \NFE, {\superdit} improves sampling efficiency by over $5\times$.
\tabref{tab:nfe} shows the {\FID} scores of {\superdit} across varying NFEs and baselines:
(1) With \NFE$=50$ (\textbf{$\mathbf{5\times}$ speedup}), \superdit outperforms {\REPA} with \NFE$=250$.
(2) With \NFE only $25$ (\textbf{$\mathbf{10\times}$ speedup}), \superdit still surpasses {\SiT} with \NFE$=250$.

\subsection{Manipulating Memory for Diverse Images}
\label{sec:exp_diversity}
\paragraph{Diverse image generation.}
We first demonstrate in Figure \figref{fig:diverse} that even for memory snippets present in the training set, \superdit can generate images that deviate significantly from the training data.
This illustrates that \superdit does not merely memorize images from the training set, but instead leverages the network's learned generalization capabilities to produce novel samples.
Specifically, when a memory snippet corresponding to the semantic concept of a "lighthouse" is provided, the generated image of the lighthouse exhibits clear differences in background, color, and other attributes compared to the most similar image in the training set.
Similarly, when a memory snippet representing an "orange" is given, \superdit generates an image of an orange placed on a plate, showing distinct differences in presentation from the close-up images of oranges in the training set.

\paragraph{Novel image generation.}
The decoupled memory bank allows {\superdit} to generate entirely new classes or samples without additional training.
By integrating snippets via SVD-based methods (see \secref{sec:dmd}), {\superdit} can embed external knowledge into its memory bank and synthesize previously unseen categories.
For instance, as shown in \figref{fig:novel}, providing a ``low-poly'' style snippet lets {\superdit} produce  images exhibit ``low-poly'' characteristics but distinct from the original input—even if such a style was never in the training data.
This training-free style transfer highlights how {\superdit} can leverage memory snippets to create novel samples and underscores the flexibility of its decoupled memory bank in adapting to new styles or categories without retraining.

\paragraph{Compositional image generation.}
Next, we will explore generating new concepts by manipulating existing snippets through latent space interpolation.
Rather than introducing a new snippet, we linearly combine two snippets stored in \(\mathbf{M}\) and the interpolated snippet produces images blending features from both ``parent'' snippets, creating hybrid or novel concepts.
For example, interpolating between a ``dog'' and a ``hat'' snippet generates a new concept ``dog wearing a hat''.
This approach also allows introducing artistic styles to existing classes, e.g., we can generate a ``swan swimming in a green river'' by interpolating between a internal snippet ``swan'' and a external knowledge ``green background'' snippet.
These results highlight \superdit's ability to composite internal and external knowledge for generating new concepts.
\looseness=-1

\subsection{Ablation Study}

The performance of \superdit depends on several factors: bank size, backbone architecture, solver, and masking strategies.
We conduct ablation studies on ImageNet $256 \times 256$ with epochs = $64$, and found that while each factor slightly impacts optimal performance, \superdit consistently generates high-quality images efficiently.

\paragraph{Optimal masking ratio for memory bank.}
\label{sec:ablation_masking}
We investigate the impact of varying masking ratio on model performance.
As illustrated in \tabref{tab:ablation_other}, a moderate mask ratio of $0.4$ balances performance and generalization, reducing reliance on memory snippets while avoiding performance degradation from excessive masking.
This configuration achieves the best performance, yielding an \FID$=5.70$.

We also evaluate different masking strategies outlined in \appref{app:memory_bank_masking_solver}.
\superdit adapts effectively to all three strategies, demonstrating robustness to noisy memory bank conditions.
\looseness=-1

\paragraph{SVD-based decomposing strategy enhances generalization.}
\label{sec:ablation_svd}
We investigate the impact of memory bank compression methods, with results summarized in \tabref{tab:ablation_other}.
Applying the decomposing strategy from \secref{sec:dmd} while halving the memory bank size reduces {\FID} by approximately $0.15$, likely due to increased sample diversity from slight noise.
Additionally, we explore snippet-level compression by randomly selecting $640$ samples per class, resulting in a half-sized memory bank.
This reduction degrades {\FID} performance by only $0.02$, highlighting substantial redundancy in the memory bank and underscoring the necessity of compression.
\looseness=-1

\paragraph{Additional findings.} We show that SDE solver is always better than ODE solver in \appref{app:memory_bank_masking_solver} and \tabref{tab:ablation_other}.
We also find that larger backbones show better performance and \superdit generalize well with various backbones and visual tokenizers in \appref{app:backbone_tokenizer} and \tabref{tab:architectures}.

\section{Conclusion}
\label{sec:conlusions}
In this paper, we introduce a new paradigm for diffusion models, \superdit, which accelerates both training and sampling processes by decoupling the memorization from the model backbone into a memory bank.
{\superdit} achieves SoTA performance on CIFAR-10 and ImageNet $256\times256$ and $512\times512$ and improving training efficiency by more than 50$\times$ and sampling efficiency by over 10$\times$ on ImageNet $256\time256$.
Additionally, the decoupled memory bank supports training-free adaptation to new images not present in the training set.

\clearpage
\section*{Acknowledgements}
This work is supported in part by the Research Center for Industries of the Future (RCIF) at Westlake University, Westlake Education Foundation, and Westlake University Center for High-performance Computing.

\section*{Impact Statement}
This paper presents work whose goal is to advance the field of Machine Learning.
There are many potential societal consequences of our work, none which we feel must be specifically highlighted here.

\bibliography{paper.bbl}
\bibliographystyle{configuration/icml2025}

\newpage
\appendix
\onecolumn

We include additional analysis, experiments details, and results of downstream tasks in this supplementary material.
In Section~\ref{app:experiments}, we present additional experiment, including
the ablation of different encoders, different backbones and tokenizers, memory bank size, masking strategies, and solver.
In Section~\ref{app:impl_detail}, we offer further implementation details on the training settings of \superdit{} on the CIFAR-10 dataset, ImageNet 64$\times$64 dataset, and ImageNet 256$\times$256 dataset.
In Section~\ref{app:interpolation}, we provide detailed analysis on the interpolation experiments.
In Section~\ref{app:transfer}, we demonstrate that the bank learned from one dataset can be easily transferred to another dataset without any additional training.
Besides, in Section~\ref{app:text2image}, we also show that \superdit can be seamlessly applied to text-to-image generation tasks.
Finally, in Section~\ref{deriveation}, we supplement the derivation of the objective functions for flow-based and diffusion-based models.

\section{Supplementary Experiments}
\label{app:experiments}
In this seciton, we include a supplementary experiments that apply \superdit  to further validate the effectiveness of {\superdit} on various downstream tasks.

\subsection{Ablation Study on Vision Encoder} \label{app:encoder}

We conduct an ablation study to investigate the impact of different encoders on the performance of {\superdit}.
Specifically, we choose the visual encoder from CLIP~\citep{radford2021clip} as an alternative to show that {\superdit} can be applied across encoders.

\paragraph{Experimental Setup}
To avoid redundant computational overhead, instead of training {\superdit} from scratch, we train only a multilayer perceptron (MLP) with dimensions $768 \times 768$ to map the features output by the CLIP visual encoder to the memory bank.
We utilize the Adam optimizer with a learning rate of $1 \times 10^{-4}$ and train the MLP for one epoch on the ImageNet $256 \times 256$ dataset.

\paragraph{Results}

\begin{figure*}
    \centering
    \includegraphics[width=0.9\textwidth]{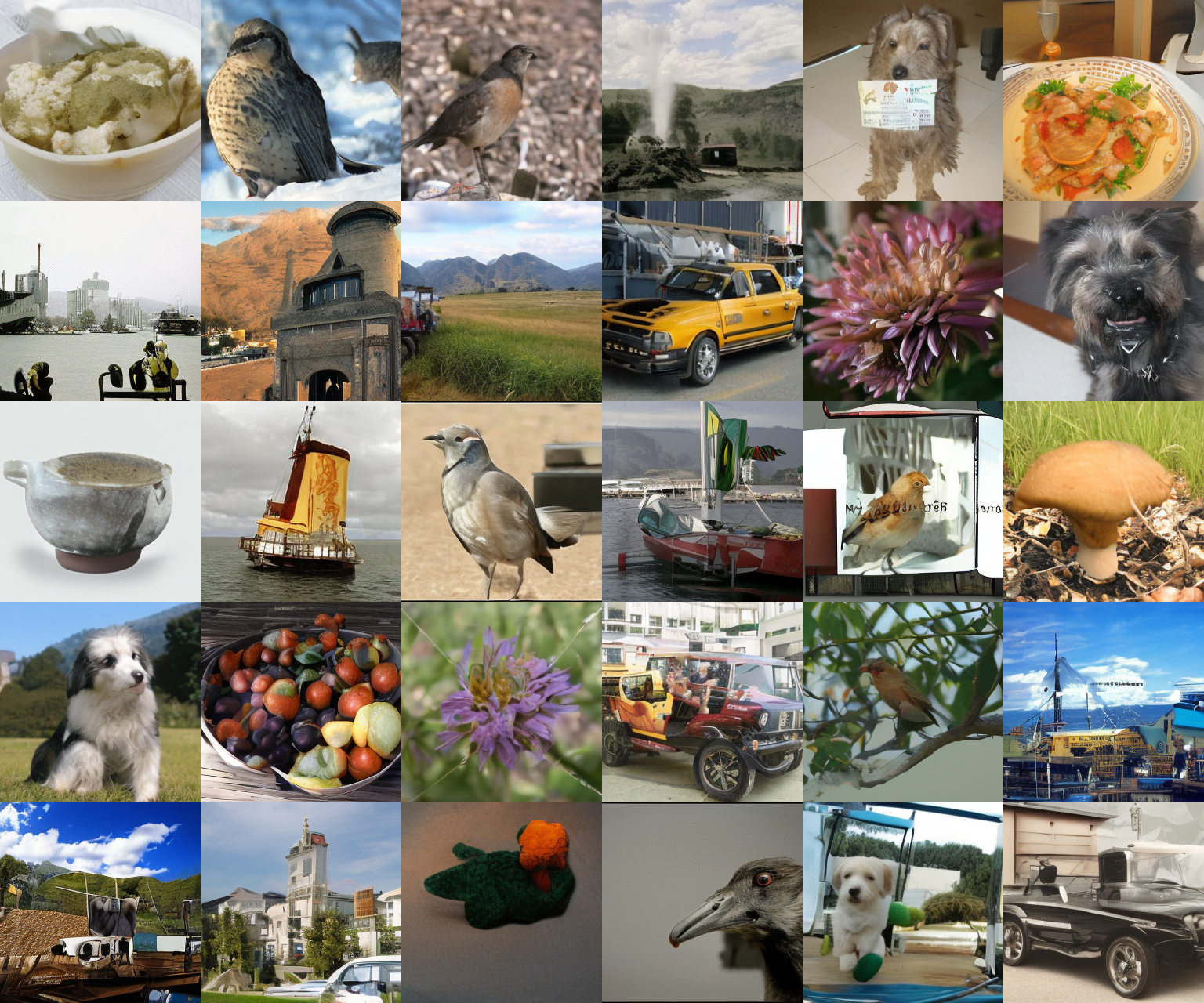}
    \caption{ \textbf{Ablation study on the encoder}: Image generation using the CLIP visual encoder with {\superdit} applied to SiT-XL/2, trained on the ImageNet $256 \times 256$ dataset.} \label{fig:appendix:encoder}
\end{figure*}

We present the results of the ablation study in Figure~\ref{fig:appendix:encoder}.

\subsection{Ablation Study on Memory Bank Size, Masking Strategies and Solver} \label{app:memory_bank_masking_solver}
\begin{table}[!ht]
    \centering\small
    \captionof{table}{\small
        \textbf{Ablation study and sensitive analysis}.
        All models use FasterDiT-B, trained for $\Epoch\!=\!64$ w/o using classifier-free guidance on ImageNet $256\times256$.
        $\downarrow$ means lower is better.
        \looseness=-1
    }
    \vspace{-0.07in}
    \resizebox{.45\textwidth}{!}{%
        \begin{tabular}{c c c c c c c}
            \toprule
            \textbf{Bank size}        & \textbf{SVD}                      & \textbf{Mask strategy}   & \textbf{Mask ratio}   & \textbf{Solver}         & \textbf{FID$\downarrow$} \\
            \midrule
            \cellcolor{orange!15}1.2B & \cellcolor{orange!15}{\checkmark} & \cellcolor{red!15}Zero   & \cellcolor{red!15}0.4 & \cellcolor{green!15}SDE & \textbf{5.70}            \\

            \midrule

            \cellcolor{orange!15}1.2B & \cellcolor{orange!15}$\times$     & Zero                     & 0.4                   & SDE                     & {5.85}                   \\
            \cellcolor{orange!15}640K & \cellcolor{orange!15}{\checkmark} & Zero                     & 0.4                   & SDE                     & {5.72}                   \\

            \midrule

            1.2B                      & {\checkmark}                      & \cellcolor{red!15}Noise  & 0.4                   & SDE                     & {6.79}                   \\
            1.2B                      & {\checkmark}                      & \cellcolor{red!15}Random & 0.4                   & SDE                     & {6.62}                   \\
            1.2B                      & {\checkmark}                      & Zero                     & \cellcolor{red!15}0   & SDE                     & {6.28}                   \\
            1.2B                      & {\checkmark}                      & Zero                     & \cellcolor{red!15}0.3 & SDE                     & {5.75}                   \\

            \midrule

            1.2B                      & {\checkmark}                      & Zero                     & 0.4                   & \cellcolor{green!15}ODE & {6.70}                   \\

            \bottomrule
        \end{tabular}%
    }
    \label{tab:ablation_other}
    \vspace{-0.05in}
\end{table}

\paragraph{Additional Masking Strategies.} We also explored two other masking strategies: \textit{random mask} and \textit{noise mask}. Specifically, \textit{random mask} replaces a randomly selected portion of each batch with Gaussian noise, while \textit{noise mask} adds noise to the entire memory snippet. The results for these two masking strategies are presented in \tabref{tab:ablation_other}. We found that zeroing out part of the snippet (the \textit{Zero mask} strategy) consistently performed best across all experiments. Therefore, we adopted \textit{Zero mask} for all major experiments.

\paragraph{SDE solver is superior.}
\label{sec:ablation_sde}
SDE solvers consistently outperform ODE solvers, reducing {\FID} by $1.0$ (\tabref{tab:ablation_other}). Thus, SDE solvers are used in all main experiments.

\subsection{Ablation Study on Model Backbones and Tokenizers} \label{app:backbone_tokenizer}

\paragraph{Larger architectures excel.}
\begin{table}[ht!]
    \centering\tiny
    \captionof{table}{\small
        \textbf{{\superdit} consistently generates high-quality samples across different backbones and image tokenizers.}
        This table reports the \FID of \superdit with different backbones and visual tokenizers on ImageNet $256\times256$.
        For a fair comparison, we train all models for 64 epochs.
        $\downarrow$ means lower is better and all results reported are without classifier-free guidance.
        \looseness=-1
    }
    \label{tab:architectures}
    \vspace{-0.10in}
    \resizebox{0.43\textwidth}{!}{%
        \begin{tabular}{lcccr}
            \toprule[0.4pt]
            \textbf{Backbone} & \textbf{Tokenizer} & \textbf{\#Params} & \textbf{Epoch} & \textbf{FID$\downarrow$} \\
            \midrule[0.2pt]
            SiT-B             & SD-VAE             & 130M              & 64             & {22.25}                  \\
            SiT-L             & SD-VAE             & 458M              & 64             & {6.49}                   \\
            SiT-XL            & SD-VAE             & 675M              & 64             & {6.31}                   \\
            \midrule[0.2pt]
            LightningDiT-B       & VA-VAE             & 130M              & 64             & {5.70}                   \\
            LightningDiT-B       & DC-AE              & 130M              & 64             & {5.97}                   \\
            \bottomrule[0.4pt]
        \end{tabular}
    }

    \vspace{-0.11in}
\end{table}

\begin{figure}[t]

	\centering
	\includegraphics[width=0.4\linewidth]{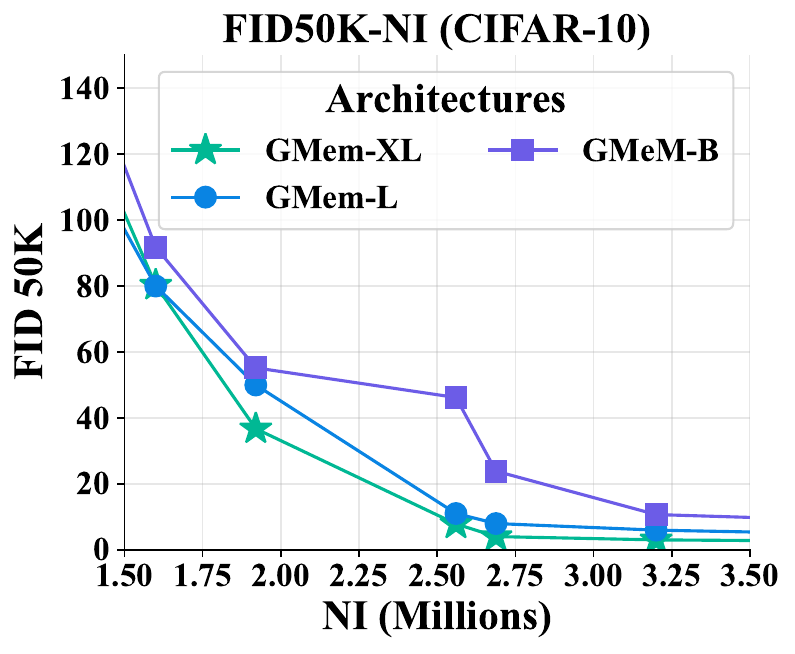} \\
	\vspace{-1em}
	\caption{\small
		\textbf{Ablation study on the variation of \FID-50K with respect to number of training images (NI). }
		We analyze the impact of NI during the training of \superdit using different backbones on CIFAR-10 dataset.
		Specifically, while \superdit-XL initially shows a higher \FID than \superdit-L at the early stages of training, after $1.75$M NI, \superdit-XL consistently outperforms both \superdit-L and \superdit-B in terms of \FID and maintains this lower value until convergence, where the final \FID reaches $1.22$.
	}
	\label{fig:architecture}
\end{figure}

We examine the scalability of \superdit by testing various model sizes and architectural configurations.
\tabref{tab:architectures} presents the \FID-50K scores of \superdit across different model sizes on ImageNet $256\times256$: larger models not only converge faster but also achieve lower \FID.
This trend aligns with findings from \citet{yu2024representation} and \citet{ma2024sit} on diffusion transformers and extends to pixel-space generation.
For example, on CIFAR-10, \superdit-XL achieves an \FID$=1.22$, outperforming smaller model variants efficiently, as depicted in~\figref{fig:architecture}.

\paragraph{\superdit generalize well with various backbones and visual tokenizers.}
We compare different backbone architectures and visual tokenizers in \figref{tab:architectures}.
LightningDiT consistently outperforms SiT under identical configurations, corroborating findings from \citet{yao2025reconstruction}.
Additionally, DC-AE and VA-VAE tokenizer yields better results than SD-VAE
, likely due to their larger parameter capacity.

\section{Implementation Details}
\label{app:impl_detail}

\subsection{Diffusion transformer architecture}

We closely follow the architecture used in \REPA~\citep{yu2024representation} and SiT~\citep{ma2024sit}.
Similar to a Vision Transformer~\citep{dosovitskiy2021an}, In this architecture, the input image is divided into patches, reshaped into a one-dimensional sequence of length $N$, and then processed by the model.
Unlike the original SiT, \REPA includes additional modulation layers called AdaIN-zero layers at each attention block.
These layers scale and shift each hidden state based on the given timestep and additional conditions.

For latent space generation, similar to \REPA, our architecture uses a downsampled latent image $z = E(x)$ as input, where $x$ is an RGB image and $E$ is the encoder of the Stable Diffusion Variational Autoencoder (VAE)~\citep{rombach2022vae}.
For pixel space generation, we remove the encoder and directly use the RGB image as input.
Specifically, we modify the original SiT by changing the number of channels from 4 to 3 and directly feed the transformed RGB image into the model.
For ImageNet $64\times64$, as an example of pixel space generation, we aslo adjust the patch size of SiT from 4 to 2 to maintain a consistent sequence length of $N=256$ patches.

Furthermore, unlike the original \REPA, instead of using class label embeddings, we utilize the CLS token from the target representation output by the encoder as an additional condition for the diffusion transformer.

\subsection{Hyperprarameters}

\paragraph{Additional implementation details}

We implement our model based on the original \REPA implementation~\citep{yu2024representation}.
We use the AdamW optimizer~\citep{kingma2014adam} with a constant learning rate of $1 \times 10^{-4}$, $\beta_1 = 0.9$, and $\beta_2 = 0.999$, without weight decay.
To accelerate training, we employ mixed-precision (fp16) computation along with gradient clipping.

For latent space generation, we pre-compute compressed latent vectors from raw images using the Stable Diffusion VAE~\citep{rombach2022vae} and utilize these latent vectors as input.
In contrast, for pixel space generation, we directly use the raw pixel data as input.
Although we experimented with data augmentations such as flipping, we found that they did not significantly improve performance.
Therefore, we do not apply any data augmentation in our experiments.

For projecting memory snippets into the backbone hidden dimension, we utilize a three-layer MLP with SiLU activations for diffusion transformers, following~\citet{yu2024representation}.
We provide detailed hyperparameter configurations in Tables~\ref{tab:appendix:setting_cifar10} and~\ref{tab:appendix:setting_in1k}.

\begin{table*}[t]
    \centering
    \caption{\small
        \textbf{Training settings of CIFAR-10.}
        We provide the training settings for all models and training algorithms on the CIFAR-10 dataset.
    }
    \label{tab:appendix:setting_cifar10}
    \small{
        \begin{tabular}{|l|ccc|} %
            \hline %
                                                                            & \multicolumn{3}{c|}{Model Size}                                                 \\
                                                                            & B                               & L                     & XL                    \\
            \hline

            \multicolumn{1}{|l@{}}{\textbf{Model details}}                  &                                 &                       &                       \\
            \hline
            Batch size                                                      & 128                             & 128                   & 128                   \\
            Training iterations                                             & 200K                            & 200k                  & 200k                  \\
            Learning rate                                                   & 1e-4                            & 1e-4                  & 1e-4                  \\
            Optimizer                                                       & Adam                            & Adam                  & Adam                  \\
            Adam $\beta_1$                                                  & 0.9                             & 0.9                   & 0.9                   \\
            Adam $\beta_2$                                                  & 0.999                           & 0.999                 & 0.999                 \\
            \hline

            \multicolumn{1}{|l@{}}{\textbf{Interpolants}}                   &                                 &                       &                       \\
            \hline
            $\alpha_t$                                                      & $1-t$                           & $1-t$                 & $1-t$                 \\
            $\sigma_t$                                                      & $t$                             & $t$                   & $t$                   \\
            $\omega_t$                                                      & $\sigma_t$                      & $\sigma_t$            & $\sigma_t$            \\
            Training Objective                                              & v-prediction                    & v-prediction          & v-prediction          \\
            Sampler                                                         & Euler                           & Euler                 & Euler                 \\
            Sampling steps                                                  & 50                              & 50                    & 50                    \\
            Classifier-free Guidance                                        & $\times$                        & $\times$              & $\times$              \\
            \hline

            \multicolumn{1}{|l@{}}{\textbf{Training details of backbone }}  &                                 &                       &                       \\ \hline
            Capacity(Mparams)                                               & 130                             & 458                   & 675                   \\
            Input dim.                                                      & 32$\times$32$\times$3           & 32$\times$32$\times$3 & 32$\times$32$\times$3 \\
            Num. layers                                                     & 12                              & 24                    & 28                    \\
            Hidden dim.                                                     & 768                             & 1,024                 & 1,152                 \\
            Num. heads                                                      & 12                              & 12                    & 16                    \\
            \hline

            \multicolumn{1}{|l@{}}{\textbf{Training details of \superdit }} &                                 &                       &                       \\ \hline
            Bank size                                                       & 50k                             & 50k                   & 50k                   \\
            Bank similarity measure                                         & Euclidean                       & Euclidean             & Euclidean             \\
            Bank objective                                                  & MSE                             & MSE                   & MSE                   \\
            Encoder $f(\xx)$                                                & DINOv2-B                        & DINOv2-B              & DINOv2-B              \\
            \hline
        \end{tabular}%
    }
\end{table*}

\begin{table*}[t]
    \centering
    \caption{\small
        \textbf{Training settings.}
        We present the training settings for all models and training algorithms on the ImageNet 256$\times$256 dataset (left) and the ImageNet 512$\times$512 dataset (right).
    }
    \label{tab:appendix:setting_in1k}
    \begin{minipage}{0.475\textwidth}
        \resizebox{\textwidth}{!}{%
            \begin{tabular}{|l|ccc|} %
                \hline %
                                                                                  & \multicolumn{3}{c|}{Model Size}                                                 \\
                                                                                  & B                               & L                     & XL                    \\
                \hline

                \multicolumn{1}{|l@{}}{\textbf{Model details}}                    &                                 &                       &                       \\
                \hline
                Batch size                                                        & 1024                             & 1024                   & 1024                   \\
                Training iterations                                               & 200K                            & 200K                  & 200K                  \\
                Learning rate                                                     & 1e-4                            & 1e-4                  & 1e-4                  \\
                Optimizer                                                         & Adam                            & Adam                  & Adam                  \\
                Adam $\beta_1$                                                    & 0.9                             & 0.9                   & 0.9                   \\
                Adam $\beta_2$                                                    & 0.999                           & 0.999                 & 0.999                 \\
                \hline

                \multicolumn{1}{|l@{}}{\textbf{Interpolants}}                     &                                 &                       &                       \\
                \hline
                $\alpha_t$                                                        & $1-t$                           & $1-t$                 & $1-t$                 \\
                $\sigma_t$                                                        & $t$                             & $t$                   & $t$                   \\
                $\omega_t$                                                        & $\sigma_t$                      & $\sigma_t$            & $\sigma_t$            \\
                Training Objective                                                & v-prediction                    & v-prediction          & v-prediction          \\
                Sampler                                                           & Heun                            & Heun                  & Heun                 \\ 
                Sampling steps                                                    & 100                             & 100                    & 100                    \\
                \hline

                \multicolumn{1}{|l@{}}{\textbf{Training details of backbone }}    &                                 &                       &                       \\ \hline
                Capacity (Mparams)                                                & 130                             & 458                   & 675                   \\
                Input dim.                                                        & 32$\times$32$\times$3           & 32$\times$32$\times$3 & 64$\times$64$\times$3 \\
                Num. layers                                                       & 12                              & 24                    & 28                    \\
                Hidden dim.                                                       & 768                             & 1,024                 & 1,152                 \\
                Num. heads                                                        & 12                              & 12                    & 16                    \\
                \hline

                \multicolumn{1}{|l@{}}{\textbf{Training details of {\superdit} }} &                                 &                       &                       \\ \hline
                Bank size                                                         & 1.2M                            & 1.2M                  & 1.2M                  \\
                Encoder $f(\xx)$                                                  & DINOv2-B                        & DINOv2-B              & DINOv2-B              \\
                \hline

            \end{tabular}%
        }
    \end{minipage}
    \begin{minipage}{0.475\textwidth}
        \resizebox{\textwidth}{!}{%
            \begin{tabular}{|l|ccc|} %
                \hline %
                                                                                & \multicolumn{3}{c|}{Model Size}                                                 \\
                                                                                & B                               & L                     & XL                    \\
                \hline

                \multicolumn{1}{|l@{}}{\textbf{Model details}}                  &                                 &                       &                       \\
                \hline
                Batch size                                                      & 1024                             & 1024                   & 1024                   \\
                Training iterations                                             & 500K                            & 500k                  & 500k                  \\
                Learning rate                                                   & 1e-4                            & 1e-4                  & 1e-4                  \\
                Optimizer                                                       & Adam                            & Adam                  & Adam                  \\
                Adam $\beta_1$                                                  & 0.9                             & 0.9                   & 0.9                   \\
                Adam $\beta_2$                                                  & 0.999                           & 0.999                 & 0.999                 \\
                \hline

                \multicolumn{1}{|l@{}}{\textbf{Interpolants}}                   &                                 &                       &                       \\
                \hline
                $\alpha_t$                                                      & $1-t$                           & $1-t$                 & $1-t$                 \\
                $\sigma_t$                                                      & $t$                             & $t$                   & $t$                   \\
                $\omega_t$                                                      & $\sigma_t$                      & $\sigma_t$            & $\sigma_t$            \\
                Training Objective                                              & v-prediction                    & v-prediction          & v-prediction          \\
                Sampler                                                         & Heun                           & Heun                 & Heun                 \\
                Sampling steps                                                  & 100                            & 100                    & 100                    \\
                \hline

                \multicolumn{1}{|l@{}}{\textbf{Training details of backbone }}  &                                 &                       &                       \\ \hline
                Capacity(Mparams)                                               & 130                             & 458                   & 675                   \\
                Input dim.                                                      & 32$\times$32$\times$3           & 32$\times$32$\times$3 & 32$\times$32$\times$3 \\
                Num. layers                                                     & 12                              & 24                    & 28                    \\
                Hidden dim.                                                     & 768                             & 1,024                 & 1,152                 \\
                Num. heads                                                      & 12                              & 12                    & 16                    \\
                \hline

                \multicolumn{1}{|l@{}}{\textbf{Training details of \superdit }} &                                 &                       &                       \\ \hline
                Bank size                                                       & 1.28M                             & 1.28M                   & 1.28M                   \\
                Encoder $f(\xx)$                                                & DINOv2-B                        & DINOv2-B              & DINOv2-B              \\
                \hline
            \end{tabular}%
        }
    \end{minipage}
\end{table*}

\paragraph{Encoder} We use Dinov2-B~\citep{oquab2023dinov2} as the encoder across all experiments, as it has been shown to significantly enhance the learning of better representations in diffusion models~\citep{yu2024representation}.
Dinov2-B offers superior performance, making it an ideal choice for facilitating the efficient training for constructing the memory bank.

\paragraph{Memory Bank} For the memory bank size, we set it to 50K for CIFAR-10 and 1.2M for ImageNet 256$\times$256 and ImageNet 512$\times$512.
Note it is essential to maintain a relatively large memory bank for optimal performance, as demonstrated in Section~\ref{sec:ablation_svd}.

\paragraph{Computing Resources} All models are primarily trained on NVIDIA H800 8-GPU setups, each equipped with 80GB of memory.
The training speed for \superdit-XL is approximately 2.71 seconds per step.

\section{Interpolation}
\label{app:interpolation}

\subsection{{\superdit} is \textbf{Not} a Memory Machine}

In this section, we provide further evidence that {\superdit} is not merely a memory machine.
Specifically, we demonstrate that {\superdit} can generate high-quality samples even for a memory snippet $\hat{\mathbf{s}}$ that never appeared in the training set.

\subsection{Interpolation Between Memory Snippets}

To investigate this ability, we conduct an interpolation experiment.
We use the ImageNet $256 \times 256$ dataset and employ the model checkpoint obtained after 140 epochs, as described in~\tabref{tab:appendix:setting_in1k}.

We randomly select two memory snippets \(\mathbf{s}_1$ and $\mathbf{s}_2\) from the memory bank $\mathbf{M}$.
We then create nine interpolated snippets $\hat{\mathbf{s}}_i$ by linearly interpolating between $\mathbf{s}_1$ and $\mathbf{s}_2$ with interpolation coefficients $\alpha_i$ ranging from 0.1 to 0.9 in increments of 0.1.
The interpolated snippets are defined as:
\[
    \hat{\mathbf{s}}_{i} = (1 - \alpha_i) \mathbf{s}_1 + \alpha_i \mathbf{s}_2, \quad \alpha_i = 0.1 i, \quad i = 1, 2, \ldots, 9.
\]
Each interpolated memory snippet $\hat{\mathbf{s}}_i$ is then fed into the transformer block to generate images.

\subsection{Interpolation results}

\begin{figure*}
    \centering
    \includegraphics[width=0.9\textwidth]{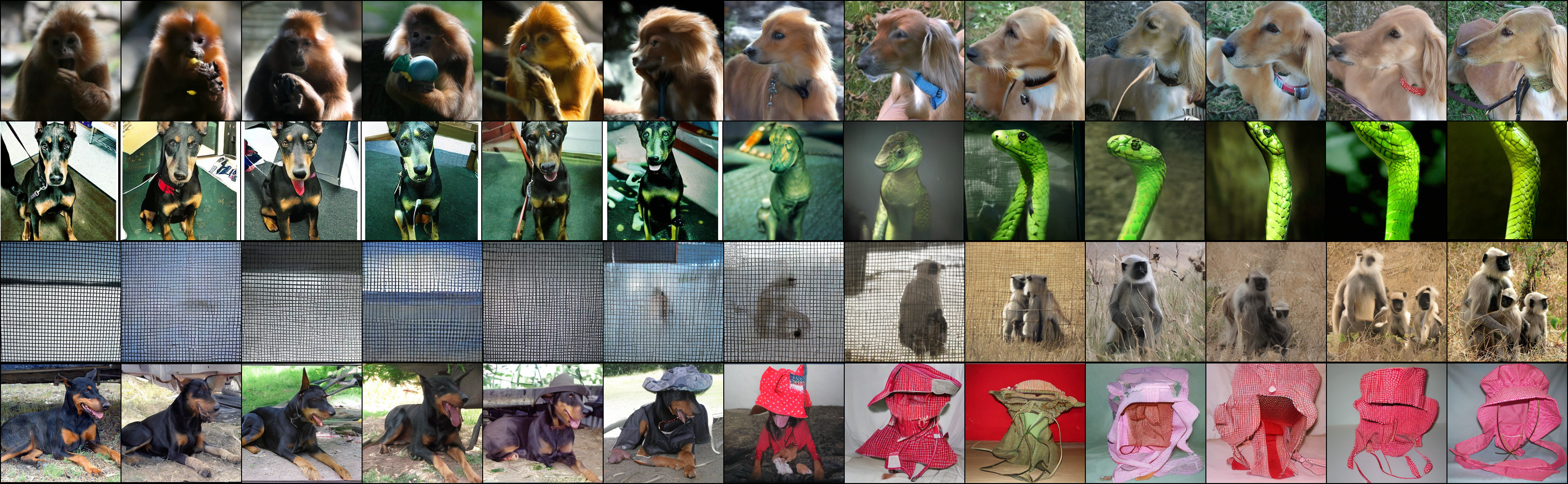}
    \caption{
        Interpolation between memory snippets. The first and last columns show the original memory snippets $\mathbf{s}_1$ and $\mathbf{s}_2$, respectively.
        The remaining columns show the generated images from the interpolated memory snippets $\hat{\mathbf{s}}_i$.}
    \label{fig:appendix:interpolation}
\end{figure*}

\begin{figure*}
    \centering
    \includegraphics[width=0.9\textwidth]{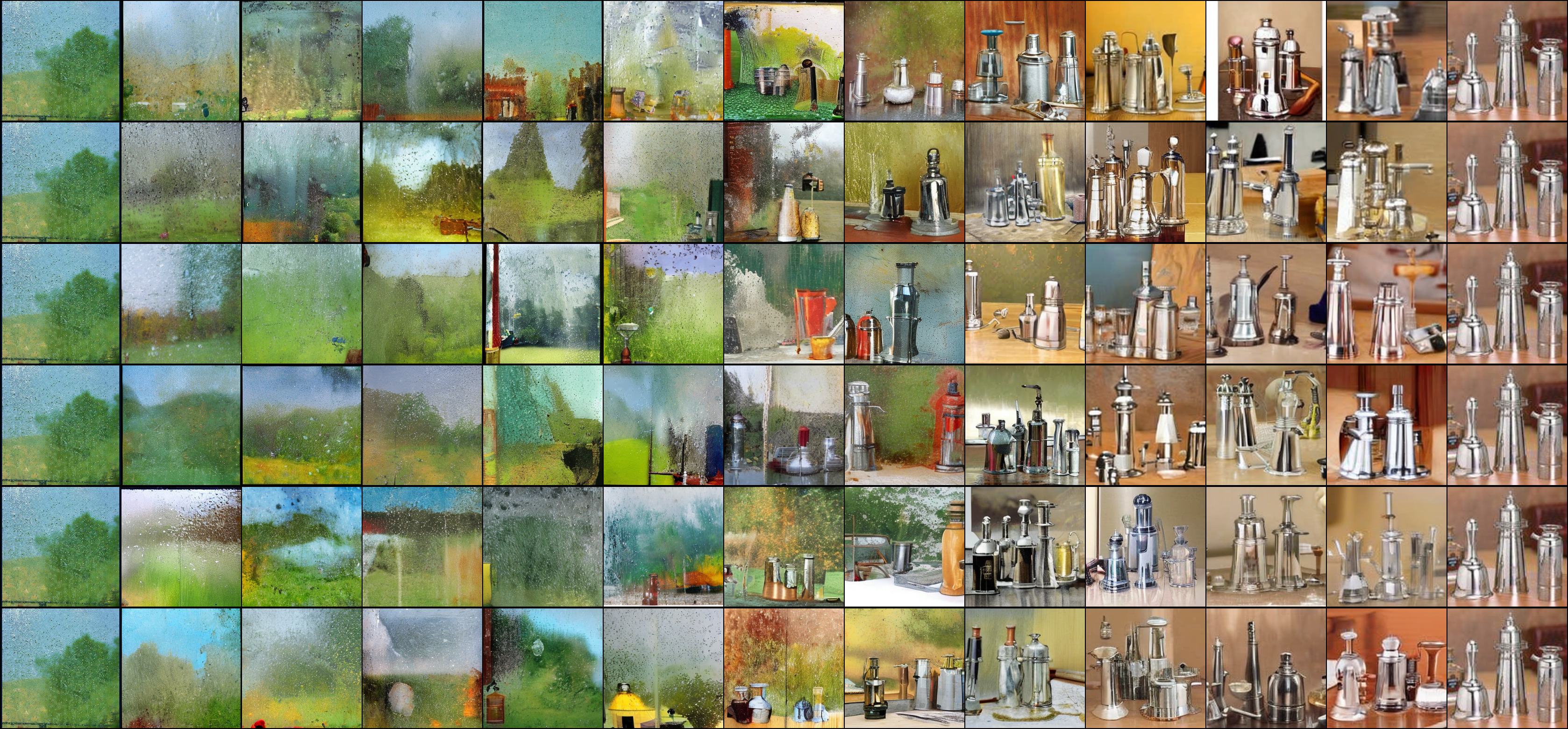}
    \caption{
        A more elaborate interpolation experiment. The first and last columns show the original memory snippets $\mathbf{s}_1$ and $\mathbf{s}_2$, respectively.
        The remaining columns show the generated images from the interpolated memory snippets $\hat{\mathbf{s}}_i$.
        Different row stands for different noise applied when generating the images.}
    \label{fig:appendix:noise}
\end{figure*}

The results of this interpolation experiment are presented in \figref{fig:appendix:interpolation} and~\figref{fig:appendix:noise}.
We observe that the generated images from the interpolated memory snippets $\hat{\mathbf{s}}_i$ are of high quality and exhibit smooth transitions between the two original memory snippets $\mathbf{s}_1$ and $\mathbf{s}_2$.

In the first row of~\figref{fig:appendix:interpolation}, we interpolate between an ape and a dog.
The dog's face gradually transforms into a smoother visage, adapting to resemble the ape.
This demonstrates representation space learned by \superdit is semantically smooth.
Surprisingly, in the second row, interpolating between a green snake and a long-faced dog results in a green reptilian creature that resembles both the snake and the dog.
This indicates that when the model encounters unseen memory snippets, it can utilize the smooth latent space to generate images similar to those it has previously encountered.

The third and last rows showcase even more imaginative interpolations.
Interpolating between a monkey and barbed wire results in an image of a monkey in a cage, while a dog and a red hat can be interpolated into a dog with a black gentleman's hat.
These outcomes suggest that the similarities captured by the model are not limited to visual resemblance but also encompass more abstract semantic similarities in the latent space.

We believe that this semantic similarity arises because our memory bank introduces additional semantic information, enabling the model to better understand the content of images.
Consequently, the model generates images that align more closely with human intuition, rather than merely memorizing the images corresponding to each snippet.

\subsection{Additional Applications}

In this section, we present experimental results demonstrating that the Memory Bank employed by {\superdit} exhibits both cross-dataset transferability and adaptability to downstream text-to-image (T2I) tasks.

\paragraph{Transferable memory bank across datasets.}
\appref{app:transfer} demonstrates the transferability and generalization of the memory bank used to guide \superdit across different datasets.
Specifically, we train $\superdit_{\textrm{IN64}}$ and $\superdit_{\textrm{CIFAR}}$ models on ImageNet $64\times64$ and CIFAR-10, respectively, resulting in memory banks of $\mathbf{M}_{\textrm{IN64}}$ and $\mathbf{M}_{\textrm{CIFAR}}$.
We then directly apply $\mathbf{M}_{\textrm{IN64}}$ to guide the sampling process of $\superdit_{\textrm{CIFAR}}$.
Our results show that $\superdit_{\textrm{CIFAR}}$ is still able to generate information consistent with the knowledge provided by $\mathbf{M}_{\textrm{IN64}}$.

\paragraph{Application to T2I generation.}
We further demonstrate that {\superdit} can be effectively applied to T2I generation tasks by encoding text prompts into memory snippets $\hat{\mathbf{s}}$, which are subsequently utilized by {\superdit} to generate corresponding images.

This integration is achieved through a two-step process:
\begin{enumerate}[label=(\alph*), nosep, leftmargin=16pt]
  \item {Pretraining the mapping function}: We pretrain a mapping function $\mpsi$ that transforms text prompt distributions $\pp({\text{text}})$ into memory snippet distributions $\pp(\mathbf{s})$ using a simple contrastive loss~\citep{radford2021clip}.
  \item {Image generation}: We then employ the pretrained mapping function $\mpsi$ in conjunction with the trained \superdit model $\mtheta$ to generate images.
\end{enumerate}

Comprehensive descriptions of the T2I generation process, experimental setup, and the resulting generated images are provided in \appref{app:transfer}.

\section{Transferability of the Memory Bank}
\label{app:transfer}

In this section, we demonstrate the transferability of the Memory Bank across different models.
Specifically, we show that the Memory Bank can be transferred between {\superdit} models trained on different datasets.
While applying a Memory Bank extracted from low-resolution images to high-resolution models (e.g., Latent Diffusion Models) may result in decreased image sharpness due to information bottlenecks, it can still enhance the diversity of the generated image.

\subsection{Experimental Setup}

To investigate the transferability, we trained a Memory Bank $\mathbf{M}_\mathrm{CIFAR}$ on the CIFAR-10 dataset and directly transferred it to a model trained on ImageNet 256 $\times$ 256 ${\superdit}_{\mathrm{IN256}}$ to guide image generation.
We used the checkpoint from the ImageNet model at 140 epochs for generation.
The detailed experimental settings are provided in~\tabref{tab:appendix:setting_cifar10} and~\tabref{tab:appendix:setting_in1k}.

\subsection{Results}

\begin{figure*}
    \centering
    \includegraphics[width=0.9\textwidth]{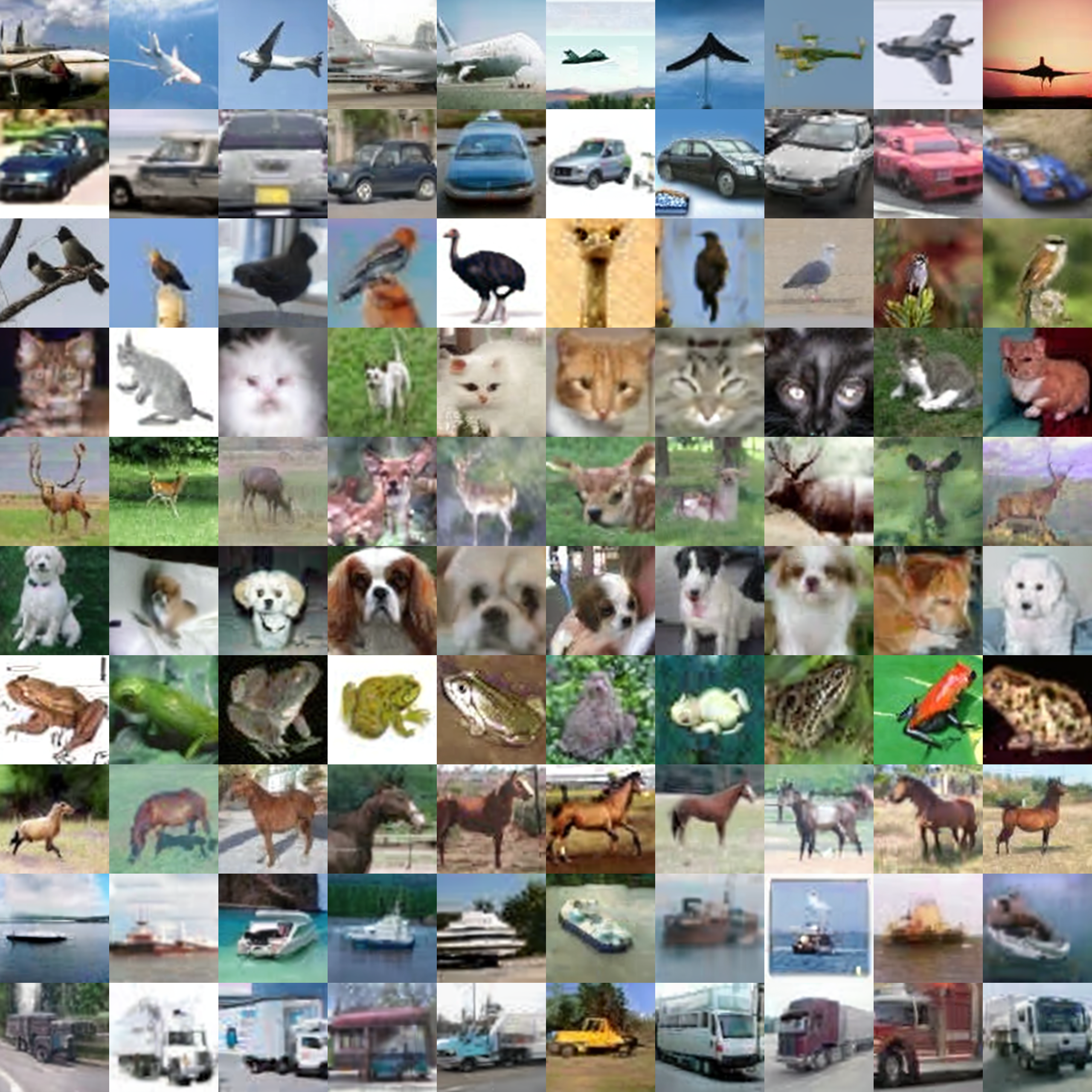}
    \caption{
        Transferability of the Memory Bank. Each row corresponding to a specifical class in CIFAR-10. Specifically, the class is from top to bottom: airplane, automobile, bird, cat, deer, dog, frog, horse, ship, and truck.
    }
    \label{fig:appendix:transfer}
\end{figure*}

\figref{fig:appendix:transfer} presents the generation results of our method on the ImageNet $256 \times 256$ dataset.
The images demonstrate that the transferred Memory Bank can effectively guide the high-resolution model. Though sharpness is limited due to information bottlenecks in memory snippets, it can improving the diversity of the generated images.

\section{Text-to-Image Generation}
\label{app:text2image}
We propose that {\superdit} can be seamlessly applied to text-to-image generation tasks.
This extension only requires training a mapping from textual CLS features to the Memory Bank, which can be efficiently implemented using a simple multilayer perceptron (MLP).

During inference, we extract CLS features from the Text Encoder using class names as input.
These features are mapped to Memory Bank snippets through a trained MLP, and then fed into the SiT backbone to generate the corresponding images.
This method enables straightforward and effective generation of images from textual descriptions.

\subsection{Experimental Setup}

In our text-to-image experiments, we utilize BERT-base~\citep{devlin2018bert} as the text encoder, using class names as textual supervision signals and the CLS token as the textual feature representation.
We employ the model trained on CIFAR-10 with 450 epochs as the generator.

For mapping text features to the Memory Bank, we use a simple MLP with dimensions 768$\times$768.
We train this mapping using the Adam optimizer with a learning rate of $1 \times 10^{-4}$, optimizing the CLIP Loss~\citep{radford2021clip} over 60 epochs.

\subsection{Experimental Results}

\begin{figure}
    \centering
    \includegraphics[width=0.47\textwidth]{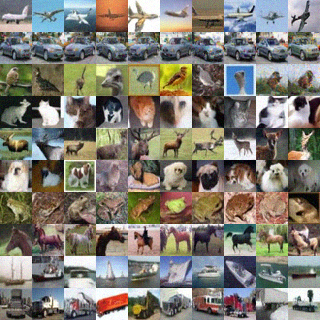}
    \caption{
        Text-to-image generation results. Each row corresponds to a specific class in CIFAR-10. Specifically, the classes are from top to bottom: airplane, automobile, bird, cat, deer, dog, frog, horse, ship, and truck.
    }
    \label{fig:appendix:t2i}
\end{figure}

We present the generated images in \figref{fig:appendix:t2i}. The results demonstrate that our method can effectively generate images that correspond closely to the provided textual descriptions.

\section{Derivation}
\label{deriveation}

In this paper, we delve into two types of generative models that learn the target distribution by training variants of denoising autoencoders: denoising diffusion probabilistic models (DDPM) and stochastic interpolants.

\subsection{Denoising Diffusion Probabilistic Models}

Diffusion models~\citep{ho2020denoising} aim to model a target distribution \( p(\xx) \) by learning a gradual denoising process that transitions from a Gaussian distribution \( \cN(0, \mI) \) to \( p(\xx) \).
The core idea is to learn the reverse process \( p(\xx_{t-1}|\xx_t) \) of a predefined forward process \( q(\xx_t|\xx_0) \), which incrementally adds Gaussian noise to the data starting from \( \xx_0 \sim p(\xx) \) over \( T \) time steps.

The forward process \( q(\xx_t|\xx_{t-1}) \) is defined as:
\[
    q(\xx_t|\xx_{t-1}) = \cN\left(\xx_t; \sqrt{1 - \beta_t} \, \xx_0, \beta_t^2 \mI\right),
\]
where \( \beta_t \in (0, 1) \) are small, predefined hyperparameters.

In the DDPM framework introduced by \citet{ho2020denoising}, the reverse process \( p(\xx_{t-1}|\xx_t) \) is parameterized as:

\[
    \begin{aligned}
        p(\xx_{t-1} & |\xx_t) =                                                                                                                                                      \\
        \cN\Bigg(                %
                    & \xx_{t-1}; \frac{1}{\sqrt{\alpha_t}} \left( \xx_t - \frac{\beta_t}{\sqrt{1 - \bar{\alpha}_t}} \, \varepsilon_\theta(\xx_t, t) \right), \Sigma_\theta(\xx_t, t)
        \Bigg)
    \end{aligned}
\]
where \( \alpha_t = 1 - \beta_t \), \( \bar{\alpha}_t = \prod_{i=1}^t \alpha_i \), \( \varepsilon_\theta(\xx_t, t) \) is a neural network parameterized by \( \theta \) and \( \Sigma_\theta(\xx_t, t) \) represents the learned variance.

The model is trained using a simple denoising autoencoder objective:
\[
    L_{\text{simple}} = \mathbb{E}_{\xx_0, \varepsilon, t} \left[ \left\| \varepsilon - \varepsilon_\theta(\xx_t, t) \right\|_2^2 \right],
\]
where \( \varepsilon \) is sampled from a standard normal distribution and \( t \) is uniformly sampled from \( \{1, \dots, T\} \).

For the variance \( \Sigma_\theta(\xx_t, t) \), \citet{ho2020denoising} initially set it to \( \sigma_t^2 I \) with \( \beta_t = \sigma_t^2 \).
However, \citet{nichol2021improved} demonstrated that performance improves when \( \Sigma_\theta(\xx_t, t) \) is learned jointly with \( \varepsilon_\theta(\xx_t, t) \). They propose optimizing the variational lower bound (VLB) objective:
\[
    L_{\text{vlb}} = \exp\left( v \log \beta_t + (1 - v) \log \tilde{\beta}_t \right),
\]
where \( v \) is a per-dimension component from the model output and \( \tilde{\beta}_t = \frac{1 - \bar{\alpha}_{t-1}}{1 - \bar{\alpha}_t} \beta_t \).

By choosing a sufficiently large \( T \) and an appropriate schedule for \( \beta_t \), the distribution \( p(\xx_T) \) approaches an isotropic Gaussian.
This allows for sample generation by starting from random noise and iteratively applying the learned reverse process \( p(\xx_{t-1}|\xx_t) \) to obtain a data sample \( \xx_0 \)~citep{ho2020denoising}.

\subsection{Stochastic interpolating}
In contrast to DDPM, flow-based models~\citep{esser2024scaling,liu2023dream} address continuous time-dependent processes involving data samples \( \xx^* \sim p(\xx) \) and Gaussian noise \( \varepsilon \sim \mathcal{N}(0, \mI) \) over the interval \( t \in [0, 1] \). The process is formulated as:

\[
    \xx_t = \alpha_t \xx_0 + \sigma_t \varepsilon, \quad \text{with} \quad \alpha_0 = \sigma_1 = 1, \quad \alpha_1 = \sigma_0 = 0,
\]

where \( \alpha_t \) decreases and \( \sigma_t \) increases as functions of \( t \).
There exists a probability flow ordinary differential equation (PF-ODE) characterized by a velocity field \( \dot{\xx}_t = \vv(\xx_t, t) \), ensuring that the distribution at time \( t \) matches the marginal \( p_t(\xx) \).

The velocity \( \vv(\xx, t) \) is expressed as a combination of two conditional expectations:

\[
    \vv(\xx, t) = \mathbb{E}[\dot{\xx}_t \mid \xx_t = \xx] = \dot{\alpha}_t \, \mathbb{E}[\xx^* \mid \xx_t = \xx] + \dot{\sigma}_t \, \mathbb{E}[\varepsilon \mid \xx_t = \xx],
\]

which can be approximated by a model \( v_\theta(\xx_t, t) \) through minimizing the training objective:

\[
    L_{\text{velocity}}(\theta) = \mathbb{E}_{\xx^*, \varepsilon, t} \left[ \left\| v_\theta(\xx_t, t) - \dot{\alpha}_t \xx^* - \dot{\sigma}_t \varepsilon \right\|^2 \right].
\]

This approach aligns with the reverse stochastic differential equation (SDE):

\[
    d\xx_t = \vv(\xx_t, t) \, dt - \frac{1}{2} w_t s(\xx_t, t) \, dt + \sqrt{w_t} \, d \bar{\mW}_t,
\]

where the score function \( s(\xx_t, t) \) is similarly defined as:

\[
    s(\xx_t, t) = -\frac{1}{\sigma_t} \, \mathbb{E}[\varepsilon \mid \xx_t = \xx].
\]

To approximate \( s(\xx_t, t) \), one can use a model \( s_\theta(\xx_t, t) \) with the training objective:

\[
    L_{\text{score}}(\theta) = \mathbb{E}_{\xx^*, \varepsilon, t} \left[ \left\| \sigma_t s_\theta(\xx_t, t) + \varepsilon \right\|^2 \right].
\]

Since \( s(\xx, t) \) can be directly computed from \( \vv(\xx, t) \) for \( t > 0 \) using the relation:

\[
    s(\xx, t) = \frac{1}{\sigma_t} \cdot \frac{\alpha_t \vv(\xx, t) - \dot{\alpha}_t \xx}{\dot{\alpha}_t \sigma_t - \alpha_t \dot{\sigma}_t},
\]

it is sufficient to estimate either the velocity \( \vv(\xx, t) \) or the score \( s(\xx, t) \).

According to \citet{albergo2023stochastic}, stochastic interpolants satisfy the following conditions when \( \alpha_t \) and \( \sigma_t \) are chosen such that:
1. \( \alpha_t^2 + \sigma_t^2 > 0 \) for all \( t \in [0, 1] \),
2. Both \( \alpha_t \) and \( \sigma_t \) are differentiable over the interval \( [0, 1] \),
3. Boundary conditions are met: \( \alpha_1 = \sigma_0 = 0 \) and \( \alpha_0 = \sigma_1 = 1 \).

These conditions ensure an unbiased interpolation between \( \xx_0 \) and \( \varepsilon \). Consequently, simple interpolants can be utilized by defining \( \alpha_t \) and \( \sigma_t \) as straightforward functions during training and inference.
Examples include linear interpolants with \( \alpha_t = 1 - t \) and \( \sigma_t = t \), or variance-preserving (VP) interpolants with \( \alpha_t = \cos\left( \frac{\pi}{2} t \right) \) and \( \sigma_t = \sin \left( \frac{\pi}{2} t \right) \).

An additional advantage of stochastic interpolants is that the diffusion coefficient \( w_t \) remains independent when training either the score or velocity models.
This independence allows \( w_t \) to be explicitly chosen after training during the sampling phase using the reverse SDE.

It's noteworthy that existing score-based diffusion models, including DDPM~\citep{ho2020denoising}, can be interpreted within an SDE framework.
Specifically, their forward diffusion processes can be viewed as predefined (discretized) forward SDEs that converge to an equilibrium distribution \( \mathcal{N}(0, \mI) \) as \( t \to \infty \).
Training is conducted over \( [0, T] \) with a sufficiently large \( T \) (e.g., \( T = 1000 \)) to ensure that \( p(\xx_T) \) approximates an isotropic Gaussian.
Generation involves solving the corresponding reverse SDE, starting from random Gaussian noise \( \xx_T \sim \mathcal{N}(0, \mI) \).
In this context, \( \alpha_t \), \( \sigma_t \), and the diffusion coefficient \( w_t \) are implicitly defined by the forward diffusion process, potentially leading to a complex design space in score-based diffusion models~\citep{karras2022elucidating}.

\end{document}